\documentclass{article}
\usepackage{cseg_arxiv}

\usepackage{amsmath,amsfonts,bm}

\def\eqref#1{equation~\ref{#1}}

\def\1{\bm{1}}

\DeclareMathAlphabet{\mathsfit}{\encodingdefault}{\sfdefault}{m}{sl}
\SetMathAlphabet{\mathsfit}{bold}{\encodingdefault}{\sfdefault}{bx}{n}

\usepackage{amsmath}
\usepackage{amssymb}
\usepackage{booktabs}
\usepackage{array}
\usepackage{graphicx}
\usepackage{float}
\usepackage{pifont}
\usepackage{placeins}
\usepackage[table]{xcolor}
\usepackage{hyperref}
\usepackage{url}
\usepackage{textcomp}
\usepackage{tcolorbox}
\tcbuselibrary{breakable,skins}

\newcommand{\fullmark}{\textcolor{green!50!black}{\ding{51}}}
\newcommand{\nomark}{\textcolor{red!75!black}{\ding{55}}}
\newcommand{\partialmark}{%
  \begingroup
  \normalsize
  \settowidth{\dimen0}{\ensuremath{\blacktriangle}}%
  \textcolor{yellow!55!orange}{\ensuremath{\blacktriangle}}%
  \llap{\raisebox{0.00ex}{\makebox[\dimen0][c]{\textcolor{black}{\tiny\bfseries !}}}}%
  \endgroup}
\newcommand{\methodname}{CSEG}
\newcommand{\benchname}{CSEG-Bench}
\newcommand{\rmname}{CSEG-RM}

\title{Benchmarking and Training Reward Models\\
for Cross-Source Evidence Grounding in Search Agents}

\author{%
\bfseries
Haochen Zhao\textsuperscript{1}\thanks{Work done during internship at Tencent.},
Minsheng Hao\textsuperscript{2},
Zheng Li\textsuperscript{1},
Yifan Li\textsuperscript{3},
Zikang Liu\textsuperscript{3},
Bowen Qin\textsuperscript{2},\\[0.4ex]
Yuhao Cheng\textsuperscript{2},
Chuang Fan\textsuperscript{2},
Haosheng Zou\textsuperscript{2},
Jeff Chen\textsuperscript{2}\thanks{Project lead.},
Sujian Li\textsuperscript{1}\thanks{Corresponding author.}\\[1ex]
\mdseries
\textsuperscript{1}Peking University \quad
\textsuperscript{2}Tencent \quad
\textsuperscript{3}Renmin University of China
}
\begin{document}

\maketitle

\begin{abstract}
Existing approaches to training and evaluating search agents focus heavily on final-answer quality, 
largely neglecting how generated answers map back to retrieved evidence. 
Recent methods incorporate evidence citation tracking to enforce grounding 
but citation support alone does not establish source independence or whether an answer appropriately handles conflicting evidence. 
In this paper, we argue that credibility in open-web retrieval depends on
principled cross-source evidence grounding that accounts for source
independence and systematically attributes and adjudicates conflicting
evidence, while hierarchical judgments of these grounding
decisions provide a unified signal for training and evaluation. 
To this end, we propose an evaluation framework, Cross-Source Evidence Grounding (\methodname{}), 
that employs a fine-grained, six-layer decision tree over the complete search trajectory. 
We introduce \benchname{} and trainable reward models to provide targeted supervision following this framework. 
Our method offers the key advantage of eliminating the severe data skew of natural trajectories by providing targeted coverage over complex, deep-layer decision paths, such as cross-source conflict resolution and nuanced attribution, that standard rollouts fail to capture.
Specifically, (1)~we define the hierarchy at label, node, and path
levels; (2)~we construct 1,160 verified trajectories over nine canonical
paths by rewriting answers on real retrieved document pools, with independent frontier
judges reproducing the entire gold path and human experts deciding inclusion; 
(3)~we train \rmname{}, which reaches 83.9\% exact accuracy on our benchmark while lying on the performance--token efficiency Pareto frontier.
Experiments that use this
reward alone to supervise a frozen-retrieval answer policy raise the
independently judged mean label from 2.83 to 4.19, and show that learning
the hierarchy generalizes across layers at a fraction of frontier-judge
cost.
\end{abstract}

\begin{center}
\small
\href{https://github.com/HaochenZhao/CSEG-Open}{%
  \raisebox{-0.25em}{\includegraphics[height=1.4em]{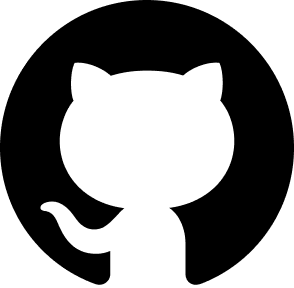}}}%
: CSEG Public Codebase\\[0.4ex]
\href{https://huggingface.co/collections/VectorZhao/cseg-rm}{%
  \raisebox{-0.25em}{\includegraphics[height=1.4em]{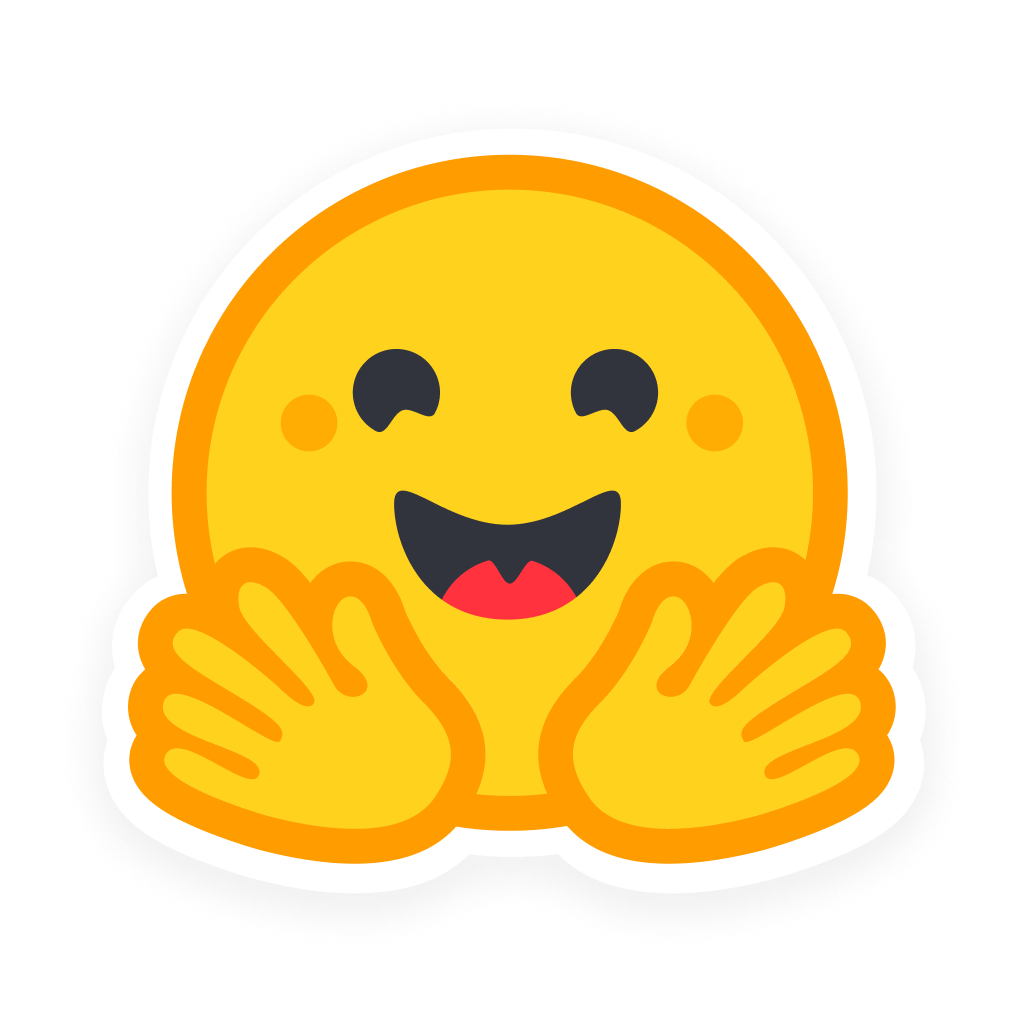}}}%
: CSEG-Bench \& CSEG-RM
\end{center}

\section{Introduction}

\begin{figure}[!ht]
\centering
\includegraphics[width=\linewidth]{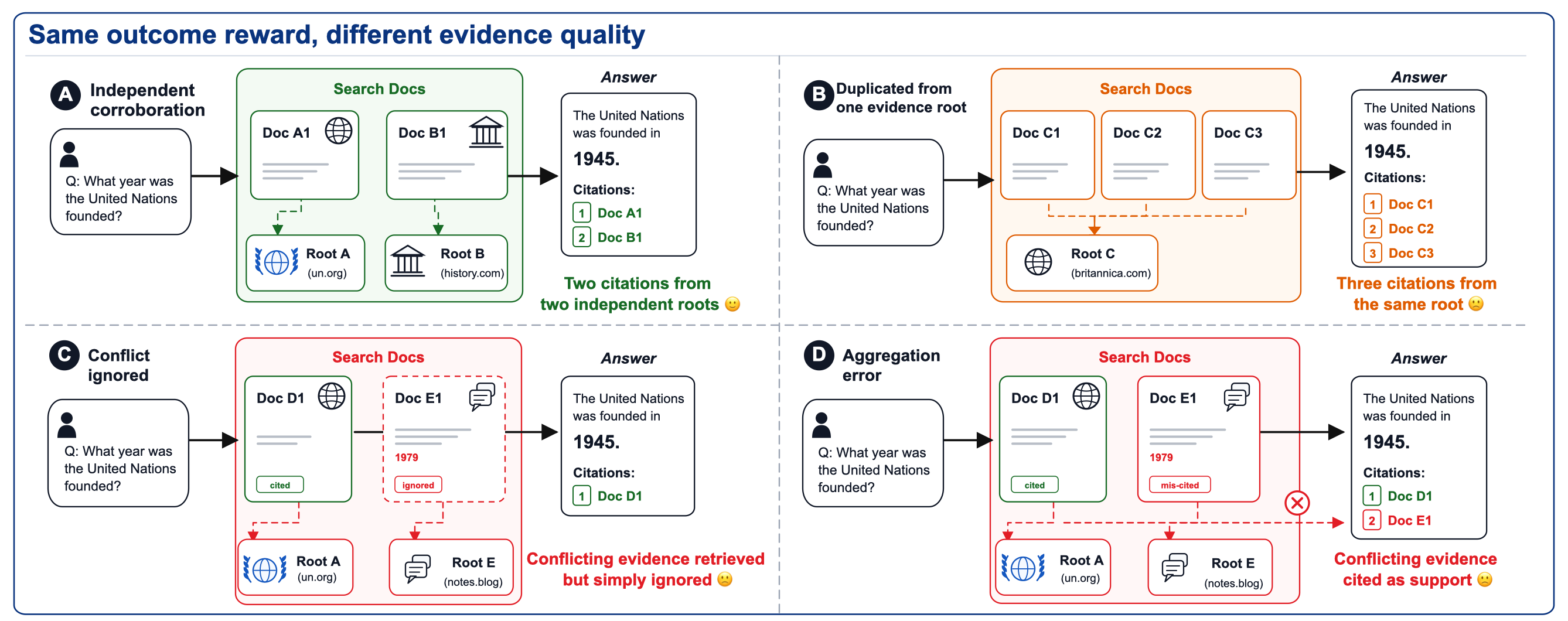}
\caption{Same answer, different evidence quality. All four trajectories answer
1945 and receive outcome reward 1, yet their evidence differs:
(A)~two citations from independent evidence roots;
(B)~three citations from the same evidence root;
(C)~conflicting evidence is retrieved but ignored; and
(D)~conflicting evidence is incorrectly cited as support.
An outcome-only reward cannot distinguish these trajectories.}
\label{fig:overview}
\end{figure}

Search agents are increasingly used for open-web information seeking and specialized applications in finance, medicine, and scientific research \citep{xi2025deepsearchsurvey}. A key capability in these settings is to retrieve and integrate evidence from multiple sources, especially when useful information is distributed across documents. However, current search agents can produce right answers while relying on wrong evidence. 
They may treat dependent sources as independent support, or integrate only one side of a noisy open-web pool in which documents both corroborate and contradict one another.
Figure~\ref{fig:overview} illustrates this right-answer, wrong-evidence problem.

Existing search agents often overlook this problem. Search-R1, R1-Searcher, and ReSearch mainly supervise the final outcome, reducing the retrieve--read--answer trajectory to a scalar reward from a rule-based verifier \citep{jin2025searchr1,song2025r1searcher,chen2025research}. As a result, trajectories with very different evidence quality can receive the same reward. Citation-aware and attribution-aware methods instead check whether a citation supports a claim \citep{gao2023alce,yue2023attrscore,huang2024citationrewards,chuang2025selfcite}. These objectives assess evidential support but do not jointly evaluate source independence and conflict handling. In open-web search, evidence is often spread across multiple sources that may differ or even conflict. Existing objectives therefore do not fully measure whether an agent has selected and integrated the right evidence across sources.

This gap also appears in real search trajectories. Our preliminary experiment shows that, in 256 production trajectories from the Tencent Yuanbao search assistant, all of whose answers already pass a closed-QA outcome verifier, 33.2\% cite multiple URLs that trace back to the same evidence root. Such false corroboration would still pass citation-based metrics, despite providing no independent support. This motivates \emph{cross-source evidence grounding}: evaluating whether sources provide independent evidence, whether they conflict, and how such conflicts should be resolved based on source authority and recency. The approaches compared in Table~\ref{tab:related-comparison} do not jointly evaluate these decisions over complete search trajectories and provide a corresponding trainable grounding reward. A useful judgment should therefore be trajectory-conditional, root-aware, and conflict-sensitive, while producing verdicts that can serve as benchmark metrics, distillation targets, and training rewards.

To address this, we propose \methodname{}, a six-layer decision tree over the complete search
trajectory---question, retrieved pool, answer, and citation links---that returns an
ordered label from 1 to 7 and the path that produced it. We design a data synthesis pipeline based on \methodname{} that yields \benchname{}, comprising 1,160 verified
trajectories. By jointly evaluating source independence, conflict handling, and
adjudication over complete search trajectories, \benchname{} addresses gaps in
existing evaluation frameworks (Table~\ref{tab:related-comparison}). We evaluate
major open-weight model families on this benchmark and examine how model size
and thinking effort affect CSEG judgment (Table~\ref{tab:main-results}). 

Further, we train Qwen3-4B and Qwen3-8B on training data consisting of 17,477 samples to see whether a compact reward model can reliably learn the grounding hierarchy. The 4B \rmname{} reaches 83.9\% exact accuracy and lies on the accuracy--cost Pareto frontier among open-weight judges. We then test whether this learned grounding signal transfers to search-agent training. On 512 frozen-retrieval prefixes, \rmname{} raises the independently judged mean label from 2.83 to 4.19, and the proportion of trajectories with labels 5--7 from 19.1\% to 40.2\%.

We contribute (i) a verifiable hierarchical measure for benchmarking cross-source
evidence grounding at label, node, and path levels; (ii) \benchname{}  with expert-adjudicated
path-level validation; and (iii) 4B/8B \rmname{} models trained on separate training data, together with a downstream
intervention showing that the trained reward raises a search agent's independently
judged grounding score.

\begin{table}[t]
\caption{Construct-level comparison of grounding evaluation and reward resources.
\fullmark: explicit; \partialmark: partial; \nomark: not targeted.
Symbols indicate task coverage, not latent capability.
Appendix~\ref{app:related} defines the columns and discusses each family.}
\label{tab:related-comparison}
\centering
\scriptsize
\setlength{\tabcolsep}{6pt}
\renewcommand{\arraystretch}{1.35}
\begin{tabular}{>{\raggedright\arraybackslash}p{2.00in} *{7}{c}}
\toprule
Representative works & \shortstack{Claim\\support} & \shortstack{Retrieved\\pool} & \shortstack{Root\\indep.} & Conflict & Authority & Traj. & \shortstack{RM\\resource} \\
\midrule
GAIA, BrowseComp & \nomark & \nomark & \nomark & \nomark & \nomark & \nomark & \nomark \\
Search-R1, R1-Searcher, ReSearch & \nomark & \partialmark & \nomark & \nomark & \nomark & \partialmark & \nomark \\
ALCE, AttrScore, SelfCite & \fullmark & \nomark & \nomark & \nomark & \nomark & \partialmark & \partialmark \\
RAGAS, ARES, RAGChecker & \fullmark & \partialmark & \nomark & \partialmark & \nomark & \partialmark & \partialmark \\
ClashEval, WikiContradict, RAMDocs, \emph{et al.} & \partialmark & \fullmark & \nomark & \fullmark & \partialmark & \partialmark & \nomark \\
DRAGged into Conflicts & \fullmark & \fullmark & \nomark & \fullmark & \partialmark & \fullmark & \nomark \\
AuthorityBench, ConflictRAG & \partialmark & \fullmark & \nomark & \fullmark & \fullmark & \partialmark & \partialmark \\
Prometheus, JudgeLM, RewardBench & \nomark & \nomark & \nomark & \nomark & \nomark & \nomark & \fullmark \\
\textbf{\methodname{} (ours)} & \fullmark & \fullmark & \fullmark & \fullmark & \fullmark & \fullmark & \fullmark \\
\bottomrule
\end{tabular}
\end{table}

\section{\methodname{} Evaluation Framework}
\label{sec:framework}

We define cross-source evidence grounding, operationalize it through a six-layer
hierarchy, and specify evaluation metrics at the label, node, and path levels.

\subsection{Concept Formulation}

Cross-source grounding is a relation among an answer, its citations, and the
retrieved pool, not an intrinsic property of the answer text: the same wording can
be well grounded under one pool and unsupported under another.  The unit of
evaluation is therefore a complete single-turn search trajectory $\tau=(q,D,a)$,
where $q$ is the question, $D=\{d_1,\ldots,d_K\}$ the retrieved document pool with each search doc
$d_i$ carrying a title, site, URL, authority level, publication time, and content,
and $a$ the answer whose claims may carry \texttt{[citation:N]} markers pointing to
$d_N$; a judge or reward model is a map $R_{\phi}(q,D,a)\rightarrow(y,p)$ returning
the terminal label $y\in\{1,\ldots,7\}$ and the decision path $p$.  The essential
difference from an outcome verifier is what is consumed: $R_{\phi}$ reads the
relation between claims in $a$ and specific documents in $D$, whereas an outcome
verifier reads only $a$ and a reference answer.  It therefore measures
\emph{grounding quality, not factual correctness} and complements outcome
verification rather than replacing it (Appendix~\ref{app:measure}).

\subsection{The six-layer hierarchy}

\begin{figure}[t]
\centering
\includegraphics[width=\linewidth]{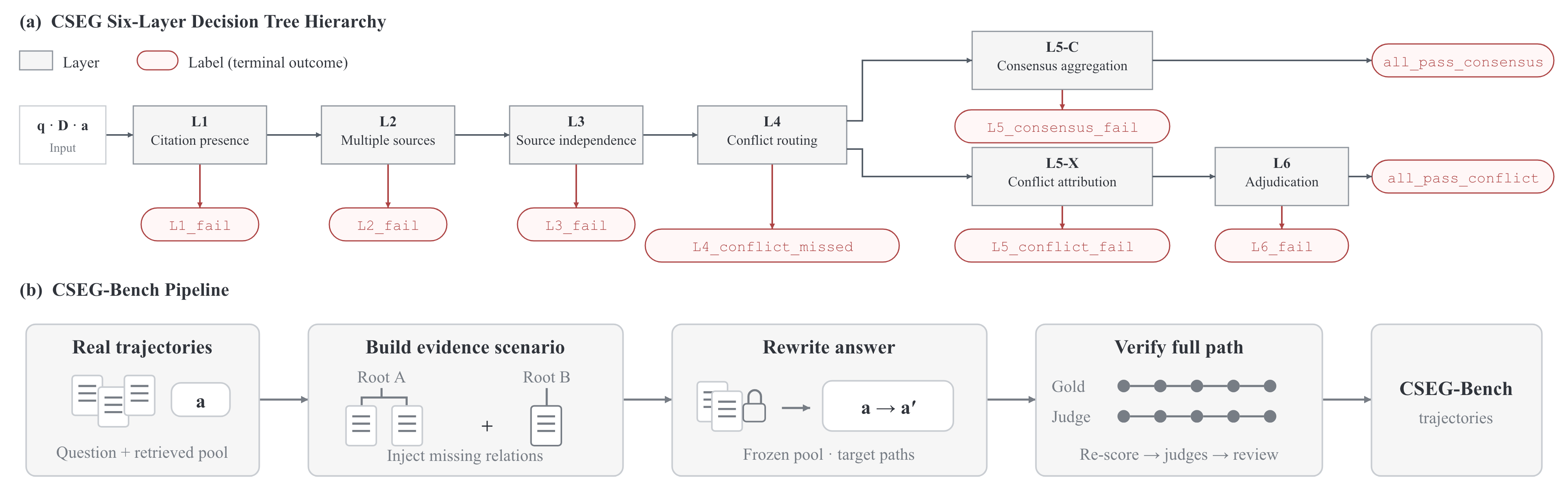}
\caption{\textbf{CSEG hierarchy and benchmark construction.}
(a) Rectangles denote layers and red capsules denote terminal outcomes. Red arrows indicate failures (labels 1--6); either all-pass outcome yields label~7.
(b) Evidence scenario construction, answer rewriting on frozen pools, and full-path
validation yield 1,160 trajectories. Roots and paths are schematic.}
\label{fig:cseg-pipelines}
\end{figure}

Figure~\ref{fig:cseg-pipelines}(a) shows the six-layer decision hierarchy of
\methodname{}, and Table~\ref{tab:hierarchy} summarizes the criterion at each layer.
Given a question, its retrieved documents, and the cited answer, the judge
checks whether a core claim has a citation to an existing, relevant document
(L1), whether it is supported by multiple cited sources (L2), and whether
those sources trace to independent evidence roots (L3).
L4 then checks for disagreement among the sources: consistent evidence
proceeds to L5-C, an identified conflict proceeds to L5-X, and a missed
conflict ends the evaluation. L5 checks how the answer combines the
evidence: L5-C (consensus aggregation) assesses whether a consensus claim has appropriate scope,
while L5-X (conflict attribution) checks whether the answer attributes the disagreement to the
sources involved. Passing L5-C terminates at \texttt{all\_pass\_consensus}.
Passing L5-X proceeds to L6, which assesses whether the answer weighs source
authority, recency, and internal consistency appropriately when adjudicating
the conflict. Passing L6 terminates at \texttt{all\_pass\_conflict}.
Evaluation stops at the first failed layer, assigning its number as the
label. Both all-pass outcomes receive label~7: the consensus branch evaluates
five layers, while the conflict branch evaluates six. L5-C and L5-X are
alternative branches of the same layer, and failure at either yields label~5.
Figure~\ref{fig:tree} in Appendix~\ref{app:measure} illustrates a complete
judgment, and Appendix~\ref{app:node-prompts} provides the node prompts,
decision criteria, and output schemas.

\begin{table}[t]
\caption{The \methodname{} hierarchy.  Labels encode the depth of the first
failed decision.}
\label{tab:hierarchy}
\centering
\footnotesize
\setlength{\tabcolsep}{3.5pt}
\begin{tabular}{c p{1.25in} p{3.65in}}
\toprule
Layer & Aspect & Criterion \\
\midrule
L1 & Citation presence & A core claim cites an existing, topically relevant document. \\
L2 & Multiple sources & At least two distinct cited sources support the core claim. \\
L3 & Source independence & Cited support traces to at least two independent evidence roots. \\
L4 & Conflict routing & Conflict classification: consistent / conflict identified / conflict missed. \\
L5-C & Consensus aggregation & Aggregation has appropriate scope and avoids over- or under-claiming. \\
L5-X & Conflict attribution & The acknowledged disagreement is attributed at source level. \\
L6 & Adjudication & The answer weighs authority, recency, and internal consistency. \\
\bottomrule
\end{tabular}
\end{table}

\subsection{Evaluation metrics at label, node, and path levels}

For comprehensive evaluation, we report \emph{label-level evaluation} (exact
accuracy, macro recall, mean absolute error (MAE), and quadratic weighted
kappa (QWK) as important metric), \emph{node-level evaluation} (each layer as its
own classifier), and \emph{path-level evaluation} (Levenshtein edit distance to the
gold path). On the ordered labels $y\in{1,\ldots,7}$, MAE is defined as
$\mathrm{MAE}=\frac{1}{n}\sum_{k=1}^{n}\lvert\hat{y}k-y_k\rvert$, while QWK is
\begin{equation}
\mathrm{QWK}
= 1-\frac{\sum{i,j=1}^{N}w_{ij}O_{ij}}{\sum_{i,j=1}^{N}w_{ij}E_{ij}},
\quad
w_{ij}=\frac{(i-j)^2}{(N-1)^2},
\label{eq}
\end{equation}
where $n$ is the number of evaluated examples, $y_k$ and $\hat{y}_k$ are
example $k$'s gold and predicted labels, and $N=7$ is the number of labels.
For gold label $i$ and predicted label $j$, $O_{ij}$ counts the observed
examples, $E_{ij}=(\sum_b O_{ib})(\sum_a O_{aj})/n$ is the expected count
under independent gold and predicted labels with the same marginals, and
$w_{ij}$ is the normalized squared distance between the labels; $a$ and $b$
sum over all $N$ labels.
Node metrics use only examples whose gold path reaches the node and count
premature stopping against recall, preventing skipped decisions from inflating scores.
We also report mean completion tokens per judgment as a cost proxy:
when the judge serves as an RM for every rollout, long generations increase
scoring latency and compute cost, potentially making reward computation a
training-throughput bottleneck.
More details can be found in Appendix~\ref{app:measure}.

\section{\benchname{}}
\label{sec:data}

The construction of \benchname{} is summarized as Figure~\ref{fig:cseg-pipelines}(b). 
We begin with \emph{outcome-passing real trajectories}, then perform
\emph{evidence scenario construction and answer rewriting} to cover the nine
canonical paths, and finally apply \emph{full-path validation} to select the
released benchmark. Unmodified real trajectories separately estimate the natural
prevalence of grounding failures; the same construction procedure on disjoint
base questions supplies reward-model training data.

\subsection{Outcome-passing real trajectories}

The real trajectories come from logged RLVR rollouts of the Tencent Yuanbao
search assistant; users whose trajectories are included opted in to research
use of their data through Yuanbao's user experience program (see the Ethics
Statement).  We retain outcome-passing records containing at
least one tool call that the service's own rubric and closed-QA verifier both pass,
so that every base trajectory \emph{already earns full marks from an outcome-only
reward}---exactly the population whose cross-source quality is in question.
Proprietary syntax is normalized to five OpenAI-style messages and one search turn,
and deduplication by query reduces 12,332 records to 1,045 unique questions, which
one fixed-seed shuffle partitions into pairwise disjoint bases for benchmark
construction, reward-model training, and answer-policy optimization
(Appendix~\ref{app:results}).

\subsection{Evidence scenario construction and answer rewriting}

Under the six-layer hierarchy, natural trajectories are extremely skewed and the
deep nodes nearly empty (Appendix~\ref{app:real}), which supports neither a
balanced comparison across layers nor deep supervision for a reward model.  We
therefore preserve the real question $q$ and use its retrieved pool $D$ as the
basis for \emph{evidence scenario construction}, with minimal controlled injection
where the target path requires an evidence relation the pool lacks. We then
freeze the resulting pool for \emph{answer rewriting} to construct $a$.
Figure~\ref{fig:cseg-pipelines}(b) summarizes the construction pipeline.
Nine construction types, one per canonical path, serve both arms,
which differ only in their base pools and in how far they are filtered.  Because
the L3 and L4 verdicts depend on the documents and not only on the answer,
construction proceeds in two stages: \emph{evidence scenario construction} identifies the
core claim, groups supporting documents by evidence root, injects whatever required
property is missing, and freezes the pool; \emph{answer rewriting} then rewrites
the final \texttt{assistant(answer)} on that frozen pool so that it lands on the
target path.  Labels are computed relative to the documents actually present, so
genuinely adding an independent corroboration produces a legitimate example.  Constructed examples must
also be indistinguishable from crawled text, or the recognizability of injected
documents leaks the L3, L4, and L6 labels outright---those documents being
precisely the branch-defining ones---so three controls apply: host-pool profile
matching, random insertion with whole-pool renumbering, and a constant injection
budget across types (Appendices~\ref{app:construct} and~\ref{app:shortcut}).

\subsection{Full-path validation}
\label{sec:validation}

Full-path validation, shown as ``Verify full path'' in
Figure~\ref{fig:cseg-pipelines}(b), checks each complete decision path.
Each of the 1,856 constructed evaluation examples is first re-scored with
the same six-layer pipeline by one frontier judge Gemini 3.1 Pro, and 1,718
(92.6\%) land on their target path.  Two \emph{other} frontier models,
GPT-5.6-Sol \citep{openai2026gpt56} and Claude Opus~5 \citep{anthropic2026opus5},
at maximum reasoning effort, then independently reproduce the \emph{entire} decision path rather than
only the terminal label, so reaching the right label through the wrong route does
not count as agreement.  Human experts then read every survivor and decide case by case
whether to keep it, discarding 332 whose reproduced verdicts they disagree with
and 226 shallow \texttt{L1\_fail} and \texttt{L2\_fail} examples that essentially
every system answers correctly.  That leaves the 1,160 trajectories released as
\benchname{}: 692 carry two-judge agreement, and 468 are single-judge examples.
Table~\ref{tab:benchmark-composition} reports the sample counts for the nine
canonical path labels. Gold reach counts for conditional node metrics are
reported in Table~\ref{tab:node-f1}. 

\begin{table}[t]
\caption{Composition of \benchname{} ($n=1{,}160$). $n$ counts trajectories
for each terminal path.}
\label{tab:benchmark-composition}
\centering
\scriptsize
\setlength{\tabcolsep}{2.5pt}
\begin{tabular}{c *{9}{c}}
\toprule
Label
& \shortstack{\texttt{L1\_}\\\texttt{fail}}
& \shortstack{\texttt{L2\_}\\\texttt{fail}}
& \shortstack{\texttt{L3\_}\\\texttt{fail}}
& \shortstack{\texttt{L4\_conflict}\\\texttt{\_missed}}
& \shortstack{\texttt{L5\_consensus}\\\texttt{\_fail}}
& \shortstack{\texttt{L5\_conflict}\\\texttt{\_fail}}
& \shortstack{\texttt{L6\_fail\_}\\\texttt{conflict}}
& \shortstack{\texttt{all\_pass\_}\\\texttt{conflict}}
& \shortstack{\texttt{all\_pass\_}\\\texttt{consensus}} \\
\midrule
$n$
& 78
& 79
& 184
& 140
& 105
& 151
& 181
& 128
& 114 \\
\bottomrule
\end{tabular}
\end{table}

\section{\rmname{}: Trainable Grounding Rewards}
\label{sec:reward-model}

We flatten constructed trajectories into layer-wise training examples, train
Qwen3-4B and Qwen3-8B \rmname{} models along three routes, and use the
resulting reward to supervise a search agent's answer policy.

\subsection{Training data}

The reward model follows the same layer-wise interface as the production-compatible trajectory scorer: invoked layer by
layer, receiving \texttt{(user\_question, retrieved\_sources, final\_answer)} and
returning that layer's judgment as JSON, after which deterministic routing
composes the terminal label and the path.  A trajectory is thus flattened into one
example per visited layer---an all-pass consensus path contributes five
examples and an all-pass conflict path contributes six---each a classification
over that layer's closed vocabulary, binary everywhere except the L4 router.  Only
the \texttt{verdict} field is graded, but it is not all the model produces: the
distillation target is the teacher's complete output, a thinking block followed by
JSON carrying \texttt{verdict}, a \texttt{reason}, and layer-specific fields such
as L3's \texttt{evidence\_roots}.  Distilling the
grounds and not only the conclusion is the point of the cold start, and is what
makes the auditable fields of Section~\ref{sec:end-policy} available at all.
Training data comes from the training arm of the same pipeline---same types, same
controls, disjoint bases---in which 4,607 trajectories flatten into 17,477 layer
examples, split by base question.

\subsection{Training and evaluation}

We train each of Qwen3-4B and Qwen3-8B \citep{yang2025qwen3} along three routes,
all in thinking mode and grading only the final verdict.  \emph{SFT} distils
rejection-sampled teacher traces: Qwen3-32B answers all 17,477 layer prompts and a
trace is kept only if its verdict equals gold, its JSON parses, it was not
truncated, and it contains a thinking block, retaining 14,876 traces that pack into
14,081 training and 795 development examples.  \emph{GRPO}
\citep{shao2024deepseekmath} uses no teacher traces and instead optimizes a
verifiable reward against gold for 400 steps, $r=+1$ when the verdict equals gold
and $-1$ when the output is unparseable or off-vocabulary, with the format penalty
harsher than a wrong verdict and the KL term applied as a loss so that a format
collapse cannot be traded against it.  This stage rebuilds its data from the same
examples, balancing the training split to 488 per (layer, verdict) cell, 15 cells
and 7,320 examples; balancing is what gives rare layers---\texttt{L5\_conflict} has
only tens under natural flattening---enough within-group variance for a
group-relative advantage.
\emph{SFT+GRPO} runs the same GRPO stage from the SFT checkpoint.

Because all
three routes share data, node prompts, and router, their comparison isolates the
training route (Appendix~\ref{app:impl}). We evaluate all models on
\benchname{} using the label, node, and path metrics defined above;
Section~\ref{sec:rq3} reports the results.

\subsection{Supervising a Search Agent's Answer Policy}
\label{sec:end-policy}

We use a \methodname{} reward model as the reward
signal for the \emph{answer policy} of a search agent, on training questions
overlapping neither the benchmark nor reward-model training.  The setting is
deliberately narrow so that any effect is attributable to evidence use rather than
to a new retrieval strategy: the prefix---question, tool call, retrieved
documents---is frozen, and the policy generates
only the final \texttt{assistant(answer)}. We use Qwen3-8B in thinking mode
as the initial policy model. The reward is the
hierarchical judgment of a \rmname{} (Qwen3-4B SFT+GRPO-400 checkpoint in main experiment) called layer by layer, with
the terminal label mapped linearly onto the unit interval,
$\mathrm{score}=(y-1)/6$; no factuality reward and no reference answer enter the
objective, because the question is sharper than ``does the agent get better''---it
is whether optimizing cross-source grounding \emph{alone} moves it.

Trajectories
before and after are re-scored by an \emph{independent} frontier judge
(Gemini 3.1 Pro, \citealp{google2026gemini31pro}). 
Because grounding can be inflated as well as improved, we also partition the evaluation questions by the \emph{honest
ceiling} of their pool: where support collapses onto one evidence root no truthful
answer can exceed label 3, so trajectories above that ceiling reflect either an L3 judge error
or learned syndication-as-corroboration (Appendix~\ref{app:impl}).

\section{Experiments}
\label{sec:experiments}

Our experiments follow a progression from diagnosis to measurement and
intervention.  \textbf{RQ1} asks why synthetic data is needed for grounding
evaluation and supervision, examining false corroboration and sparse deep-layer
supervision in real trajectories and direct rollouts.  \textbf{RQ2} uses \benchname{} to identify what
factors influence cross-source evidence grounding.  \textbf{RQ3} asks whether compact, deployable models can
learn this capability, whether the resulting \rmname{} models are cost-effective,
and whether they improve a search agent's independently judged evidence use. 

\subsection{RQ1: why do we need synthetic data?}
\label{sec:rq1-skew}

We identify two key limitations of relying on natural trajectories alone: \emph{false corroboration} and
\emph{sparse deep-layer supervision}. So we need synthetic data. 
Evidence scenario construction and answer
rewriting provide controlled coverage of these grounding failures and deeper
paths, while full-path validation checks the resulting labels.
All 256 audited production trajectories pass the service's closed-QA verifier,
yet 112 (43.8\%) terminate by L3: 23 lack a valid citation, four lack
multi-source support, and 85 cite support that is not independent.  A further
32 (12.5\%) encounter a conflict without acknowledging it, and only 85 (33.2\%)
traverse the complete hierarchy.  The dominant failure is therefore
\emph{false corroboration}: several cited URLs collapse onto too few independent
evidence roots, despite receiving full credit from metrics that ask only whether
a citation is present.  These figures characterize one service, but suffice to
reject outcome success as a grounding measure (Appendix~\ref{app:real}).

On the same 256 frozen $(q,D)$ prefixes, open-weight policies that only rewrite
the final answer likewise stop in the shallow layers
(Table~\ref{tab:search-agent-zeroshot}): DeepSeek-V4-Flash
\citep{deepseekai2026deepseekv4} at low effort has mean
label 4.11 with 52.4\% terminating by L3, Qwen3-8B \citep{yang2025qwen3} in
thinking mode 2.89 / 70.1\%,
and GPT-OSS-120B \citep{openai2025gptoss} at low effort 2.37 / 84.2\%.  Those rollouts are not
outcome-filtered, so they do not reproduce the 43.8\% production-audit rate; they
show that the shallow-failure pattern is not specific to one deployed agent.

More samples do not produce balanced coverage of the hierarchy.  In the separate
Qwen3-8B thinking rollout arm used later for the training-data comparison,
flattening the natural trajectories yields 2,636 L1 examples but only 72 examples
at L5-X, compared with roughly 976 per layer in the constructed arm.
This skew suggests that direct rollouts contain too little deep-layer supervision
to train a reliable hierarchical judge.  RQ3 tests that consequence directly by
holding the GRPO procedure fixed and changing only the source of training data.

\subsection{RQ2: what governs CSEG ability on \benchname{}?}
\label{sec:rq2}

Table~\ref{tab:main-results} reports the hierarchy on the verified benchmark.
Every baseline is open-weight, spanning 4B to 1.6T total parameters, and none is
the re-scoring judge or either path-reproduction judge of
Section~\ref{sec:validation}, so all rows are blind with respect to benchmark
construction.  This subsection considers the zero-shot block; the trained
\rmname{} models are analyzed in RQ3.

\paragraph{Capacity within a family.}
At fixed thinking effort, larger models within the same family are usually
stronger.  Qwen3 at high effort rises from 0.383 to 0.423 and 0.523 across 4B,
8B, and 32B; GPT-OSS at low effort rises from 0.494 at 20B to 0.596 at 120B.
The trend is not universal---GPT-OSS-120B at medium effort is slightly below
GPT-OSS-20B---so capacity is a common advantage rather than a general ranking
principle.  Across families, moreover, parameter count does not order
performance, so CSEG is not solved by scale alone.

\begin{figure}[t]
\centering
\includegraphics[width=\linewidth]{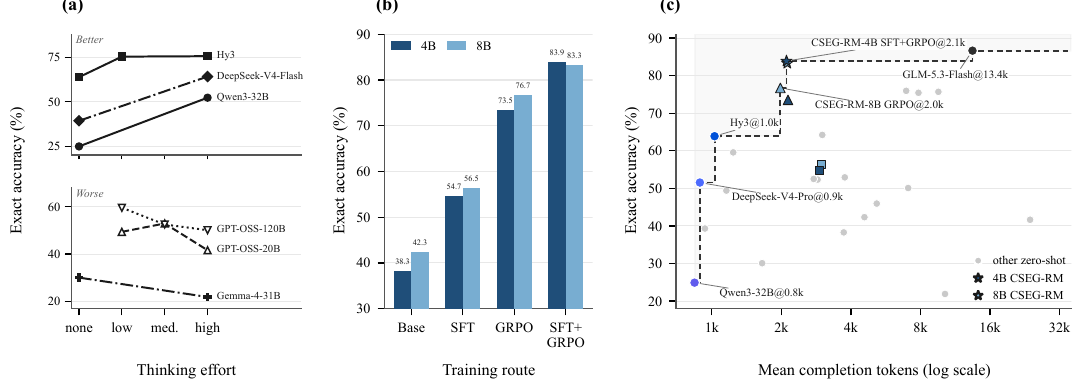}
\caption{Thinking effort, training, and accuracy--cost trade-offs on
\benchname{} ($n=1{,}160$). (a) Selected zero-shot models improve (top) or
regress (bottom) at higher thinking effort; the two plots use different
vertical scales. (b) Qwen3-4B/8B accuracy across training routes.
(c) Accuracy versus mean completion tokens; the dashed staircase is the Pareto
frontier and gray points are other zero-shot configurations. Dark/light blue
mark 4B/8B \rmname{}; squares, triangles, and stars denote SFT, GRPO, and SFT+GRPO.}
\label{fig:acc-tokens}
\end{figure}

\paragraph{Thinking effort.}
Thinking effort interacts strongly with the model family.  Qwen3-32B,
DeepSeek-V4-Flash, and DeepSeek-V4-Pro benefit substantially from additional
deliberation, whereas Hy3 \citep{tencent2026hy3} largely saturates after low
effort.  In contrast,
GPT-OSS-20B loses 11.3 exact-accuracy points from medium to high effort while
using 6.3$\times$ more tokens; Gemma-4-31B \citep{gemmateam2026gemma4} at high
effort falls 8.2 points below
its no-thinking result while using 6.2$\times$ more tokens; and GPT-OSS-120B
declines monotonically from low to high effort.  Longer deliberation over a
decision tree a model has not internalized can let it talk itself past the layer
boundaries.  Neither parameter scaling nor test-time reasoning scaling alone
reliably produces CSEG ability; Figure~\ref{fig:acc-tokens}(a) summarizes these
model-dependent trajectories.

\begin{table}[t]
\caption{Benchmark results on \benchname{} ($n=1{,}160$).  
``Total'' is the overall parameter count; ``Tokens'' is mean completion tokens per
judgment. Bold marks the best value per block. 
$\dagger$ is also the rejection-sampling SFT teacher.}
\label{tab:main-results}
\centering
\scriptsize
\setlength{\tabcolsep}{3pt}
\renewcommand{\arraystretch}{0.95}
\begin{tabular}{>{\raggedright\arraybackslash}p{1.16in}rlcccccr}
\toprule
Model & Total & Think & Exact & Macro R. & MAE $\downarrow$ & QWK $\uparrow$ & Path ED $\downarrow$ & Tokens \\
\midrule
Qwen3-4B & 4.0B & high & 0.3828 & 0.3934 & 1.5560 & 0.3590 & 1.7612 & 3,725 \\
Qwen3-8B & 8.2B & high & 0.4233 & 0.4543 & 1.3793 & 0.4809 & 1.5767 & 4,578 \\
GPT-OSS-20B & 20.9B & low & 0.4940 & 0.5454 & 1.1845 & 0.5035 & 1.5078 & 1,156 \\
GPT-OSS-20B & 20.9B & medium & 0.5293 & 0.5919 & 1.2517 & 0.4656 & 1.5690 & 3,763 \\
GPT-OSS-20B & 20.9B & high & 0.4164 & 0.4998 & 1.9655 & 0.2376 & 2.1121 & 23,860 \\
Muse-Glimmer-30B & 30B & high & 0.4595 & 0.5187 & 1.8129 & 0.3035 & 2.2440 & 5,175 \\
Gemma-4-31B & 31B & none & 0.3009 & 0.3438 & 2.4509 & 0.1886 & 2.8121 & 1,651 \\
Gemma-4-31B & 31B & high & 0.2190 & 0.3022 & 3.0198 & 0.0810 & 3.3017 & 10,221 \\
Qwen3-32B$^{\dagger}$ & 32.8B & none & 0.2491 & 0.2316 & 2.0198 & 0.0883 & 2.4414 & 842 \\
Qwen3-32B$^{\dagger}$ & 32.8B & high & 0.5233 & 0.5652 & 1.0966 & 0.5859 & 1.2569 & 2,870 \\
GPT-OSS-120B & 116.8B & low & 0.5957 & 0.6551 & 1.0216 & 0.5967 & 1.2940 & 1,238 \\
GPT-OSS-120B & 116.8B & medium & 0.5250 & 0.5966 & 1.3629 & 0.4598 & 1.6966 & 2,762 \\
GPT-OSS-120B & 116.8B & high & 0.5009 & 0.5759 & 1.5474 & 0.3723 & 1.9086 & 7,078 \\
DeepSeek-V4-Flash & 284B & none & 0.3931 & 0.4247 & 1.3724 & 0.4554 & 1.8164 & 934 \\
DeepSeek-V4-Flash & 284B & high & 0.6422 & 0.6930 & 0.8086 & 0.6916 & 0.9750 & 3,007 \\
Hy3 & 295B & none & 0.6388 & 0.6356 & 0.7509 & 0.6825 & 0.9940 & 1,028 \\
Hy3 & 295B & low & 0.7543 & 0.7865 & 0.7371 & 0.6529 & 0.9371 & 7,843 \\
Hy3 & 295B & high & 0.7569 & 0.7875 & 0.7000 & 0.6680 & 0.9000 & 9,565 \\
GLM-5.3-Flash & 320B & max & \textbf{0.8664} & \textbf{0.8796} & \textbf{0.2741} & \textbf{0.8905} & \textbf{0.3491} & 13,437 \\
GLM-5.2 & 753B & none & 0.6371 & 0.6654 & 0.9078 & 0.6142 & 1.0690 & 1,057 \\
DeepSeek-V4-Pro & 1.6T & none & 0.5155 & 0.5355 & 1.0716 & 0.5582 & 1.4879 & 888 \\
DeepSeek-V4-Pro & 1.6T & max & 0.7595 & 0.7878 & 0.6388 & 0.7003 & 0.7888 & 6,939 \\
\midrule
Qwen3-4B SFT & 4.0B & think & 0.5466 & 0.5808 & 1.0371 & 0.6136 & 1.1983 & 2,931 \\
Qwen3-4B GRPO & 4.0B & think & 0.7353 & 0.7615 & 0.6672 & 0.7052 & 0.8466 & 2,136 \\
Qwen3-4B SFT+GRPO & 4.0B & think & \textbf{0.8388} & \textbf{0.8557} & \textbf{0.3871} & \textbf{0.8236} & \textbf{0.4784} & 2,105 \\
Qwen3-8B SFT & 8.2B & think & 0.5647 & 0.5990 & 0.9491 & 0.6529 & 1.1129 & 2,984 \\
Qwen3-8B GRPO & 8.2B & think & 0.7672 & 0.7957 & 0.5776 & 0.7358 & 0.7026 & 1,980 \\
Qwen3-8B SFT+GRPO & 8.2B & think & 0.8328 & 0.8475 & \textbf{0.3871} & 0.8146 & 0.4905 & 2,121 \\
\bottomrule
\end{tabular}
\end{table}

\subsection{RQ3: can compact models learn CSEG well enough to serve as rewards?}
\label{sec:rq3}

\paragraph{Learning the hierarchy.}
All three training routes improve both Qwen3-4B and Qwen3-8B on \benchname{}.
SFT raises exact accuracy from 0.383 to 0.547 at 4B and from 0.423 to 0.565 at
8B.  GRPO reaches 0.735 and 0.767, while SFT+GRPO reaches 0.839 and 0.833 at
roughly 2.1k completion tokens (Figure~\ref{fig:acc-tokens}(b)).  The two final
models differ by only 0.6 points, indicating that the training route matters
more here than doubling model size.
Neither half is dispensable: the verifiable reward says whether a verdict is
right and nothing about why, so the grounds must arrive by distillation.  The
trained models are also cheaper than their own initializations, using
1.7$\times$ and 2.3$\times$ fewer completion tokens than the same base models at
high effort, and only one of twenty-two zero-shot configurations exceeds the
best of them.

Reporting results at three evaluation levels localizes what training fixes rather than merely
decorating terminal accuracy.  Conditional macro F1 at L5-X rises from
0.410 to 0.982, while L5-C---the layer closest to a conventional
attribution judgment---remains the weakest
(Appendix~\ref{app:results}).

\paragraph{Why constructed training data is necessary.}
\label{sec:rq2-ablation}\label{sec:data-ablation}
RQ1 motivates synthetic data through false corroboration and sparse deep-layer
supervision in natural trajectories. To test whether deliberately constructed
coverage improves learning, we compare balanced synthetic data
against Qwen3-8B natural rollouts while holding the seed questions, row counts,
splits, and GRPO recipe fixed.  The rollout arm holds $q$ and $D$ byte-identical
and regenerates only the answer before both arms are labelled, flattened,
truncated to the same row counts, and split identically.  Constructed data
improves exact accuracy by 12.6 points at 4B and 20.2 points at 8B.  The widening
gap rules out the explanation that only the smaller model fails to fit the
rollout distribution: under this training protocol, the natural distribution
lacks the required learning signal.

\paragraph{Accuracy--cost trade-off.}
Among the configurations evaluated, two trained models lie on the
accuracy--token Pareto frontier.  Qwen3-8B GRPO enters the frontier at 1,980 mean
completion tokens, while Qwen3-4B SFT+GRPO holds it until GLM-5.3-Flash
\citep{glm2026glm5}, which
uses 80$\times$ more parameters and 6.4$\times$ more tokens for 2.8 additional
accuracy points.  The practical result is therefore not that CSEG is beyond
frontier models, but that it can be distilled into a reward model cheap enough
to run once per rollout during training (Figure~\ref{fig:acc-tokens}(c)).

\paragraph{Does benchmark performance transfer downstream?}
Each of 512 frozen-retrieval questions is rolled out once from the step-0 and
step-200 checkpoints and scored by an independent frontier judge with web access
disabled (Appendix~\ref{app:results}).  Using Qwen3-4B SFT+GRPO as the reward
raises the search agent's independently judged mean label from 2.834 to 4.193
and the proportion of trajectories with labels 5--7 from 19.1\% to 40.2\%
(Table~\ref{tab:end-policy-full}).  The improvement is not merely
citation-format compliance: labels 1 and 2 both decrease, while a larger proportion of trajectories reaches
the deeper layers.

\begin{table}[t]
\caption{Answer policy before and after \methodname{}-supervised optimization, on
512 frozen-retrieval questions.  Scores come from an independent frontier judge
running the same six-layer pipeline.  The reward model is Qwen3-4B SFT+GRPO-400
(grader exact 0.8388).}
\label{tab:end-policy-full}
\centering
\scriptsize
\setlength{\tabcolsep}{4pt}
\begin{tabular}{l c c}
\toprule
Quantity & Step 0 (Qwen3-8B base) & Step 200 (SFT+GRPO-400 RM) \\
\midrule
Mean label & 2.834 & \textbf{4.193} \\
Median label & 3 & 4 \\
Label histogram, 1--7 & 186/42/123/63/53/20/25 & 20/4/172/110/123/33/50 \\
$P(\text{label}\geq4)$ & 0.314 & 0.617 \\
$P(\text{label}\geq5)$ & 0.191 & \textbf{0.402} \\
$P(\text{label}=7)$ & 0.049 & 0.098 \\
\bottomrule
\end{tabular}
\end{table}

Under the same protocol, the weaker GRPO-only reward model also improves the
policy, but yields a proportion of trajectories with labels
5--7 approximately ten percentage points lower (Figure~\ref{fig:end-policy-sftgrpo} in
Appendix~\ref{app:results}).  Across these two reward models, the model with
higher \benchname{} accuracy therefore provides stronger downstream supervision,
supporting the benchmark's predictive validity.

\section{Related Work}\label{sec:related}

Real-world web evidence is complex: sources may repeat the same information, contradict one another, or contain outdated claims and errors. Reliable search agents should handle these conditions appropriately. Recent work evaluates conflict handling and source authority in retrieved evidence \citep{cattan2025dragged,yao2026authoritybench}, yet common search-agent objectives do not explicitly reward these decisions. Outcome-based reinforcement learning rewards answer correctness without distinguishing careful evidence integration from ignored conflicts \citep{jin2025searchr1,song2025r1searcher,chen2025research}. Citation-aware methods assess evidential support, but support alone establishes neither source independence nor appropriate conflict handling \citep{gao2023alce,chuang2025selfcite}.

This gap extends to evaluation. GAIA and BrowseComp test information-seeking through verifiable answers, but their answer scores do not assess duplicated evidence or overlooked conflicts \citep{mialon2023gaia,wei2025browsecomp}. Table~\ref{tab:related-comparison} shows that existing approaches address claim support, conflict, and authority to varying degrees without jointly assessing evidence-root independence. To bridge this gap, \methodname{} judges these dimensions together over the question, retrieved pool, answer, and citation links. \benchname{} tests independent support, conflict identification and attribution, and adjudication by source authority and recency; its layer-level labels also supervise grounding reward models. Appendix~\ref{app:related} provides the full discussion of related work.

\section{Conclusion}

An outcome-only reward cannot distinguish \emph{why} an answer was right, and
citation support alone cannot establish source independence or appropriate
conflict handling across a retrieved pool.  \methodname{} makes
that capability an auditable six-layer judgment at three evaluation levels; construction
is load-bearing on both sides of the measure, making the deep layers observable and
trainable; and the resulting 4B and 8B graders sit on the Pareto frontier,
one raising a search agent's independently judged mean label from 2.83 to 4.19.
Outcome success and evidence use must be measured separately, and that measurement
can itself be learned and optimized against.

\clearpage

\section*{Ethics Statement}

The source trajectories come from the Tencent Yuanbao search assistant, and
users whose trajectories are included consented to the use of their data for
research through a dedicated user experience program.  The release is designed
to remove direct identifiers and sensitive content, minimize the exposure of
query text, document data retention and access controls, and comply with the
licenses and terms governing the retrieved pages.  Documents injected during
benchmark construction exist only inside the controlled benchmark environment
and are never presented as genuine web documents outside that setting.



\FloatBarrier

\bibliography{iclr2027_conference}
\bibliographystyle{iclr2027_conference}

\appendix
\let\csegoldsection\section
\renewcommand{\section}{\clearpage\csegoldsection}

\section{Related Work in Full}
\label{app:related}

This appendix expands Section~\ref{sec:related} and closes with a construct-level comparison.

\paragraph{Comparison dimensions.}
In Table~\ref{tab:related-comparison}, ``Claim support'' denotes claim-level
evidential support; ``Retrieved pool'' requires evaluation over a real
multi-document pool, rather than a single or curated context; web access alone
does not qualify. ``Traj.'' denotes complete $(q,D,a)$ trajectories with citation
links. ``RM resource'' requires training data, a trainable reward model, and
evidence that the reward can supervise a policy.

\paragraph{Retrieval and search-agent reinforcement learning.}
Retrieval-augmented generation connects generators to explicit corpora and has
progressed from learned retrieval to self-reflection and corrective fallback
\citep{guu2020realm,lewis2020rag,asai2023selfrag,yan2024crag}, while ReAct,
Toolformer, and WebGPT make tools or browsing part of inference
\citep{yao2022react,schick2023toolformer,nakano2021webgpt}.  Recent
reinforcement-learning systems optimize multi-step search, query generation, and
long-horizon web research
\citep{jin2025searchr1,song2025r1searcher,chen2025research,
jiang2025deepretrieval,li2025searcho1,zheng2025deepresearcher,
li2025webthinker}.  Search-R1, R1-Searcher, and ReSearch primarily reward final-answer correctness; DeepRetrieval instead rewards retrieval quality. These signals do not explicitly assess source independence, conflict attribution, or evidence-based adjudication.
\methodname{} supplies the trajectory-level grounding reward that outcome
supervision misses, together with evidence that it moves an answer policy; it is not
another search algorithm.

\paragraph{Search-agent benchmarks.}
GAIA evaluates general assistant capabilities, including web browsing and tool use, while BrowseComp focuses on finding difficult information on the web \citep{mialon2023gaia,wei2025browsecomp}. Both use verifiable final answers, making task success straightforward to score. Their answer metrics do not directly evaluate the retrieved pool, citation links, source independence, or conflict handling. Table~\ref{tab:related-comparison} therefore marks these grounding dimensions as not targeted, even though solving the tasks can require extensive search.

\paragraph{Attribution and citation-aware supervision.}
Provenance-aware and long-form QA benchmarks motivate explicit evidence use
\citep{petroni2020kilt,stelmakh2022asqa,amouyal2022qampari,
rosset2024researchy}, attribution work evaluates or trains claim--passage
support \citep{bohnet2022attributedqa,gao2023alce,huang2024citationrewards,
chuang2025selfcite}, and factuality and RAG evaluators measure consistency,
atomic accuracy, hallucination, attribution, and component quality
\citep{honovich2022true,min2023factscore,wei2024longfact,chern2023factool,
niu2024ragtruth,yue2023attrscore,es2023ragas,saadfalcon2023ares,ru2024ragchecker}.
These approaches assess evidential support and retrieval quality at different levels of granularity. ALCE includes open-domain retrieval and multi-source synthesis, so this family is not restricted to small or conflict-free contexts. The remaining gap is joint evaluation of evidence-root independence, conflict attribution, and adjudication over the retrieved pool.

\paragraph{Conflicting evidence and multi-source integration.}
A substantial literature studies parametric--context disagreement, contradictory
passages, ambiguity, misinformation, source reliability, and authority
\citep{zhou2023contextfaithful,chen2022richconflict,wu2024clasheval,
hou2024wikicontradict,wang2025conflicting,wang2024astuterag,pham2024whoswho,
cattan2025dragged,yao2026authoritybench,wang2026conflictrag}.  These studies establish conflict handling as an evaluation target. In particular, DRAGged into Conflicts uses real search results and evaluates conflict types, citation grounding, and appropriate response behavior; AuthorityBench explicitly studies source authority. Our distinction is therefore not the use of real retrieved evidence or conflict evaluation alone. \methodname{} jointly evaluates these decisions with evidence-root independence through a staged judgment of the question, retrieved pool, answer, and citation links, and uses the resulting labels to train grounding reward models.

\paragraph{Reward models and structured judges.}
LLM judges and critique or reward models provide scalable evaluation but inherit
task-specific biases \citep{zheng2023llmjudge,liu2023geval,kim2023prometheus,
kim2024prometheus2,zhu2023judgelm,lambert2024rewardbench,ke2023critiquellm,
huang2024judgelimits}.  Where structured judging exists, its structure is
typically inherited from mathematical or code reasoning steps
\citep{lightman2023verify,wang2023mathshepherd,zheng2024processbench}, and no
comparable consensus exists for how to segment a retrieval judgment.  Our
segmentation is a decision tree over evidence--answer relations, not a chain over
a model's own reasoning steps, and the supervision signal it yields is a
verifiable per-layer verdict that we then use to supervise an answer policy.

\paragraph{Surface shortcuts in constructed corpora.}
Annotation artifacts, brittle heuristics, and detectable machine-written text all
create shortcuts \citep{geirhos2020shortcut,gururangan2018artifacts,
mccoy2019hans,mitchell2023detectgpt,sadasivan2023detection}.  In a constructed
retrieval corpus, detectability converts directly into label leakage, because the
inserted documents are precisely the ones that determine the branch.  Morphology,
position, and constant-budget controls address this failure mode
(Appendix~\ref{app:shortcut}).

\section{The Natural Distribution as Evidence}
\label{app:real}

This appendix supports RQ1's motivation for synthetic data by documenting false
corroboration and sparse deep-layer coverage in natural trajectories; it carries
none of the reward-model results. The original audit of 256 production
trajectories evaluated L6 on both L4 branches; its retained measurements
therefore precede the routing revision in Figure~\ref{fig:cseg-pipelines}.
That audit yields a strongly skewed distribution
(Table~\ref{tab:real-dist}).  Three readings follow.  First, 43.8\% of real
trajectories fail within the first three layers and L3 alone accounts for 33.2\%,
so the most common failure is not a missing citation but several URLs collapsing
onto one evidence root---which any metric that asks only whether a citation
exists would score as a pass.  Second, the deep nodes are almost empty, with two
L6 failures in 256 trajectories, one on each L4 branch, and the conflict branch
touched by only 21 trajectories in total (16 failures and five passes), because
cross-source integration has never been a training objective and models are
stopped in the shallow layers.  Third, therefore, measuring \emph{and} learning
the deep nodes requires controlled construction---the second half of that claim
is what Section~\ref{sec:rq2-ablation} demonstrates---and the production set
answers only the question of what natural failure looks like.

\begin{table}[h]
\caption{Gold distribution over 256 production trajectories, every one of which
passes the service's closed-QA outcome verifier. These are the original audit counts, before consensus paths bypassed L6.}
\label{tab:real-dist}
\centering
\small
\begin{tabular}{lrrl}
\toprule
Terminal type & $n$ & Share & Depth \\
\midrule
L1 fail (no valid citation) & 23 & 9.0\% & shallow \\
L2 fail (single source) & 4 & 1.6\% & shallow \\
L3 fail (multiple URLs, one evidence root) & 85 & 33.2\% & shallow \\
L4 conflict missed & 32 & 12.5\% & middle \\
L5-X fail (conflict attribution) & 16 & 6.3\% & deep \\
L5-C fail (consensus aggregation) & 9 & 3.5\% & deep \\
L6 adjudication fail (1 conflict / 1 consensus) & 2 & 0.8\% & deep \\
Complete pass (consensus / conflict) & 80 / 5 & 33.2\% & --- \\
\bottomrule
\end{tabular}
\end{table}

\begin{table}[h]
\caption{Open-weight policies rolled out on the same 256 frozen $(q,D)$ prefixes
as Table~\ref{tab:real-dist}, rewriting only the final answer.  The production
row is the outcome-passing gold audit (mean label computed from that
distribution); the other rows are not outcome-filtered and do not replace the
43.8\% / 33.2\% claims above.}
\label{tab:search-agent-zeroshot}
\centering
\small
\setlength{\tabcolsep}{6pt}
\begin{tabular}{lcc}
\toprule
Policy & Mean label & Fail by L3 \\
\midrule
Production agent (Table~\ref{tab:real-dist}) & 4.48 & 43.8\% \\
DeepSeek-V4-Flash (low) & 4.11 & 52.4\% \\
Qwen3-8B (thinking) & 2.89 & 70.1\% \\
GPT-OSS-120B (low) & 2.37 & 84.2\% \\
\bottomrule
\end{tabular}
\end{table}
\FloatBarrier

\section{The Evaluand and the Metrics}
\label{app:measure}

\paragraph{The evaluation unit.}
As a message sequence, a trajectory $\tau=(q,D,a)$ is the five-turn structure
\texttt{system}, \texttt{user}, \texttt{assistant(tool\_call)}, \texttt{tool},
\texttt{assistant(answer)}, and the object under evaluation is the map
\begin{equation}
  R_{\phi}(q,D,a)\rightarrow (y,p), \qquad y\in\{1,\ldots,7\},
\end{equation}
where $p$ is the full decision path and $y$ its terminal label.

\paragraph{Label-level evaluation.}
The output is treated as an ordered class in $\{1,\ldots,7\}$.  Alongside exact
accuracy and macro recall we report MAE and QWK as the label metrics, with QWK primary;
out-of-range predictions are clamped into $[1,7]$.

\paragraph{Node-level evaluation.}
Every layer is scored as its own classifier, taking \texttt{fail} as the positive
class at binary nodes and reporting macro scores at the three-way L4 router.  The
metrics are \emph{conditional}: a node is scored only on examples whose gold path
actually reaches it, and a prediction that stops early is recorded as ``not
reached'' and penalized in the recall of the relevant gold class.  The leaderboard
records both unconditional and conditional precision, recall, and F1, and the
conditional form is what we report, because it separates a node's intrinsic
accuracy from the appearance of accuracy created by upstream early stopping.

\paragraph{Path-level evaluation.}
A traversal is represented as a token sequence such as
\texttt{[L1:pass, L2:pass, L3:fail]}, and we report Levenshtein edit distance to
the gold path---a substitution captures a wrong verdict, an insertion or deletion
captures stopping late or early---together with the exact whole-path match rate.
Path-level evaluation is what localizes \emph{where} in the tree a judge diverges,
which the terminal label alone cannot show.

\FloatBarrier

\begin{figure}[!ht]
\centering
\includegraphics[width=0.92\linewidth]{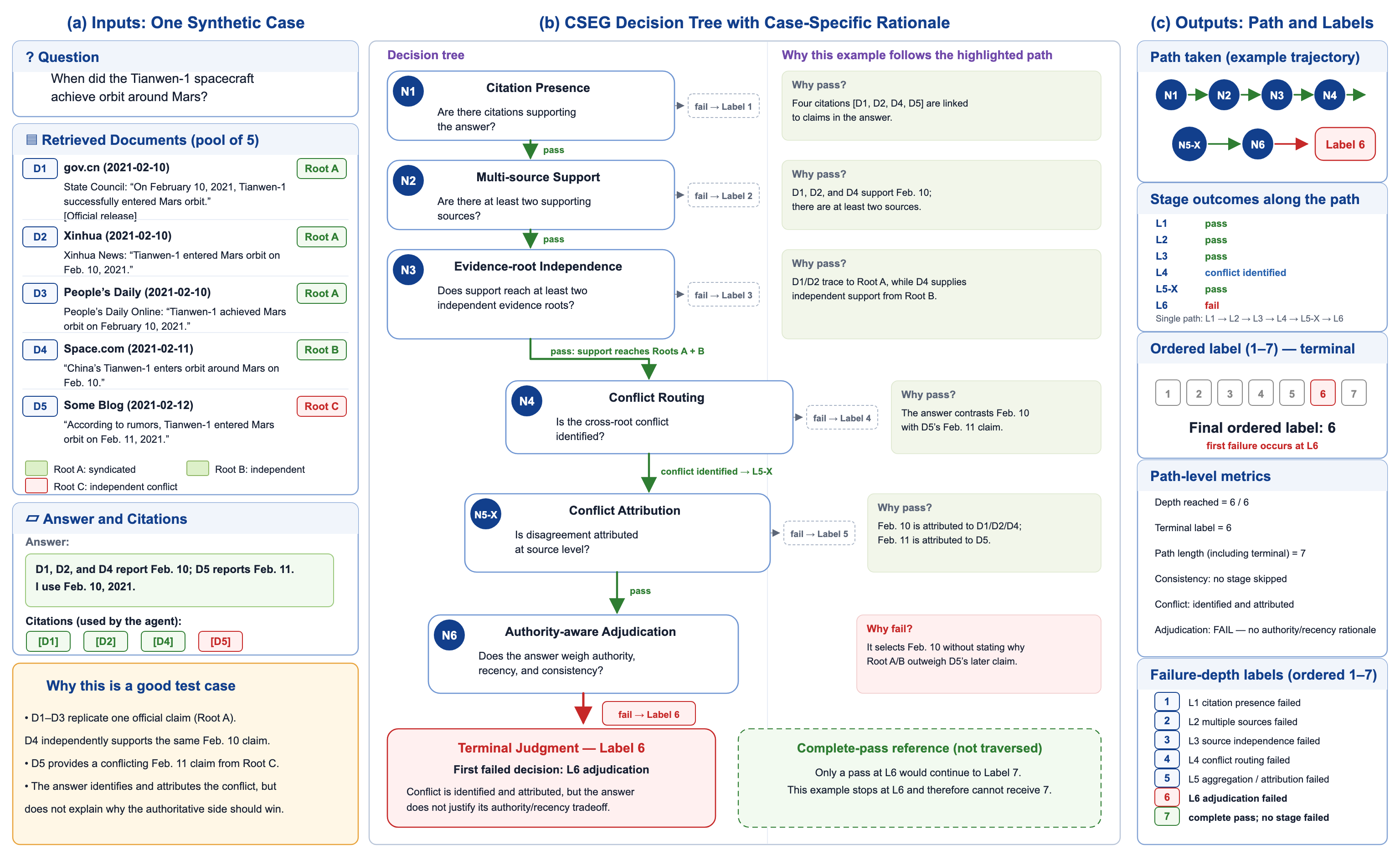}
\caption{A concrete \methodname{} judgment: inputs (left), the six layer
decisions with case-specific rationales (center), and the resulting path, node
verdicts, terminal label, and path metrics (right).}
\label{fig:tree}
\end{figure}

\paragraph{Worked example.}
Figure~\ref{fig:tree} walks one trajectory through the six-layer hierarchy: the
inputs, the layer decisions with case-specific rationales, and the resulting
path, node verdicts, terminal label, and path metrics.

\FloatBarrier

\clearpage
\paragraph{Terminology and distinctions.}
Table~\ref{tab:terminology} distinguishes contextual shorthand from terms that
refer to different objects or measurements. These distinctions apply to both the
main text and the implementation details.

\begin{table}[!ht]
\caption{Terminology guide. Related expressions are not interchangeable unless
explicitly identified as shorthand; model-specific settings retain their original
names.}
\label{tab:terminology}
\centering
\small
\setlength{\tabcolsep}{5pt}
\renewcommand{\arraystretch}{1.2}
\begin{tabular}{>{\raggedright\arraybackslash}p{1.16in}
                >{\raggedright\arraybackslash}p{1.56in}
                >{\raggedright\arraybackslash}p{2.20in}}
\toprule
Concept & Related expressions & Usage and distinction \\
\midrule
Reasoning configuration & Thinking effort; reasoning effort; deliberation &
Effort denotes a model-specific setting. Thinking mode indicates whether thinking
is enabled; deliberation describes the generated reasoning process. Matching
setting names do not establish equal reasoning budgets across models. \\
\addlinespace
Label prediction & Exact accuracy; terminal accuracy; accuracy &
In benchmark label results, these refer to exact label accuracy: agreement between
predicted and gold terminal labels. Exact whole-path match instead requires the
entire decision path to agree. \\
\addlinespace
Evaluation role & Judge; scorer; grader; reward model (RM) &
A judge makes evaluation decisions; the scorer composes layer decisions through
routing. An RM supplies a training reward. Grader is contextual shorthand for
the scoring model, not a separate model family; an independent evaluation judge
is distinct from the RM used for policy training. \\
\addlinespace
Efficiency measure & Completion tokens; token efficiency; cost &
Completion-token count is the reported cost proxy. Accuracy--token comparisons
do not by themselves measure latency, monetary cost, or compute across model
sizes. \\
\addlinespace
Hierarchy structure & Layer; node &
There are six layers and seven node classifiers because layer L5 has alternative
L5-C and L5-X nodes. The node prompts use the same layer terminology. Terminal label~7
means that every visited layer passed: five on the consensus branch or six
on the conflict branch. No seventh layer is evaluated. \\
\addlinespace
Evidence provenance & Document; source; evidence root &
A document is a retrieved item; cited source refers to a document cited as
evidence. Its evidence root is the origin of its support for the core claim.
Different documents or websites need not provide independent evidence roots. \\
\bottomrule
\end{tabular}
\end{table}
\FloatBarrier

\section{Construction and Validation Detail}
\label{app:construct}

\paragraph{The nine canonical paths.}
Table~\ref{tab:types} lists the construction types.  One set of types serves both
arms.  The evaluation arm yields 1,856 constructed trajectories and passes the
full validation loop of Section~\ref{sec:validation}; the training arm yields
4,607 trajectories on a disjoint base pool and passes only the target-path
re-scoring and a strict path check, because the path-reproduction and expert-review
stage removes exactly the hardest examples, which is the part a reward model most needs.

\begin{table}[h]
\caption{The nine canonical paths.  ``Independent,'' ``redundant,'' and
``conflict'' are defined relative to the core claim.}
\label{tab:types}
\centering
\small
\setlength{\tabcolsep}{3pt}
\begin{tabular}{l c l l}
\toprule
Type & Label & Required terminal behavior & Retrieved-pool requirement \\
\midrule
\texttt{L1\_fail} & 1 & L1 fail & --- \\
\texttt{L2\_fail} & 2 & L1 pass; L2 fail & independent \\
\texttt{L3\_fail} & 3 & L1--L2 pass; L3 fail & redundant \\
\texttt{L4\_conflict\_missed} & 4 & conflict exists but is missed & independent, conflict \\
\texttt{L5\_consensus\_fail} & 5 & consistent pool; aggregation fails & independent, no conflict \\
\texttt{L5\_conflict\_fail} & 5 & conflict identified; attribution fails & independent, conflict \\
\texttt{L6\_fail\_conflict} & 6 & attribution passes; adjudication fails & independent, conflict \\
\texttt{all\_pass\_consensus} & 7 & L5-C pass; terminate without L6 & independent, no conflict \\
\texttt{all\_pass\_conflict} & 7 & L5-X pass; L6 pass; terminate & independent, conflict \\
\bottomrule
\end{tabular}
\end{table}

\paragraph{Two-stage construction.}
\emph{Scenario construction} first identifies the core answerable claim of the
base, groups the documents that support that claim by evidence root, and derives
whether the pool offers independent support, redundancy, or conflict---the same
yardstick the L3 judge uses.  When a required property is absent, documents
carrying that property are injected; labels are computed relative to the documents
that are actually present, so genuinely adding an independent corroboration
produces a legitimate and correctly labelled example.  A base that already
violates an exclusion condition, such as a conflicted pool requested for a
consensus type, is skipped rather than repaired, because appending documents
cannot remove a property that already exists.  Edited scenarios are cached by base
and requirement signature so that answer variants can share an identical frozen
pool.  \emph{Answer rewriting} then rewrites the final
\texttt{assistant(answer)} on that frozen pool so that it lands on the target
path.  The constructor also emits a rationale, which serves only to let experts
audit whether construction hit the intended path; it is never used as a
supervision target.

\paragraph{Surface-form controls.}
Each injected document is matched to the \emph{profile of its own host pool}:
punctuation width, \texttt{publish\_time} rendering, line layout, authority
vocabulary, and crawler metadata wrappers.  Documents are inserted at uniformly
random positions and the whole pool is renumbered $1..K$; regardless of target
type every pool receives the same fixed number of additions (two by default), with
any remainder filled by documents that are topically adjacent but silent on the
core claim.  On the answer side, the placement convention for
\texttt{[citation:N]} is measured from that trajectory's own original answer.
Appendix~\ref{app:shortcut} measures what each control removes.

\section{Why Surface Form Is a Label Shortcut}
\label{app:shortcut}

The three surface-form controls of Appendix~\ref{app:construct} exist because in
a constructed retrieval corpus the recognizability of an injected document is not
a cosmetic matter but a label shortcut: the injected documents are exactly the
ones that determine L3, L4, and L6.  This appendix records the construction-stage
audit that fixed each control.

\paragraph{Layout fingerprints.}
On unaligned injected documents, the handwritten rule ``\texttt{publish\_time} is
an ISO date, the body has no crawler noise, and the body is shorter than 900
Chinese characters'' identified injected documents with 100.0\% precision and
90.5\% recall against a 9.8\% base rate---1,760 injected documents among
16,140 originals---while firing on none of the 3,328 documents in the 256
production trajectories.  Alignment is therefore performed against \emph{each
trajectory's own host pool} rather than a global template;
\texttt{authority\_level} is only collapsed onto the pool's vocabulary and
\texttt{publish\_time} only reformatted, so that the L6 signal is not erased.
After these controls the detector makes no positive predictions at all: recall and
false-positive rate are both zero and precision is undefined.

\paragraph{Position and numbering.}
Appending injections at the end of the pool leaks the label through numbering and
interacts with truncation, which then removes the decisive documents: under a
15,872-token budget, 33.8\% of edited records lost every injected document, which
uniformly random insertion with whole-pool renumbering reduced to 4.9\%; at 20,480
tokens the corresponding rates were 8.8\% and 1.4\%.  Truncation is performed at
complete document boundaries.

\paragraph{Injection count and type coupling.}
If only the missing property is supplied, pool size becomes a function of the
branch: the shallow, consensus, and conflict types add 0.00, 0.77, and 1.30
documents on average.  Pool size on its own is nearly harmless, because the
natural dispersion of pool size swamps a one-document shift---a classifier given
only the pool size gains essentially nothing on the seven-way label.  The danger
is the coupling, and it is conditional: the count is the payload and the
recognizability of injected documents is the key.  Given an oracle that
identifies which documents were planted, counting alone predicts the coarse
branch 72.9\% of the time against a 54.3\% majority baseline, and
$\text{planted}=2$ implies the conflict branch with 100\% precision; on the
seven-way label the same oracle count gives 25.3\% against a 20.0\% baseline.
Branch prediction is the dangerous half, because L4 routing is the layer every
judge is weakest at, and it is exactly what a free branch label would hand over.
Holding the planted-document budget constant across all types severs the payload
from the key: the count carries no branch information whatever the detector
does.  Filler documents are safe precisely because the whole pipeline is defined
relative to the core claim: a document that says nothing about the core claim can
neither be a second evidence root for it nor be redundant or in conflict with the
documents that support it.

\section{Design Decisions Behind the Hierarchy}
\label{app:design}

\paragraph{The responsibility boundary at L1.}
If L1 also judged full semantic support, overclaiming examples would terminate
before L5 and the deeper tree would be unreachable.  L1 is therefore restricted to
citation existence and topical relevance, and scope, overclaiming, and entailment
belong to L5.  Correspondingly, the \texttt{L1\_fail} construction specification
requires an answer with no citation, a citation to a document number that does not
exist, or a citation to an off-topic document; ``topically relevant but
insufficient support'' now passes L1 and terminates at L5.

\paragraph{Decoupling L6 from tone.}
L6 failure is defined only by substance---preferring lower-authority, staler, or
internally inconsistent evidence.  In early data roughly 81\% of L6-fail examples
carried emphatic or absolute phrasing, and the judge's stated reasons touched
both the tone and the substance, so absolute phrasing correlated with adjudication
failure strongly enough to be learnable as a shortcut in place of the capability.
Construction therefore requires neutral tone, the criterion never references
confidence of expression, and each new batch is checked for the proportion of emphatic
phrasing per label group falling back to the baseline of the other groups.

\paragraph{L3 by evidence root rather than by site.}
Defined over distinct websites, L3 passes almost always in real pools and carries
no information.  Tracing citations to evidence roots exposes duplicated reporting
and makes L3 the largest failure bucket in production traffic
(33.2\%, Appendix~\ref{app:real}).  The layer emits \texttt{evidence\_roots} and
\texttt{distinct\_root\_count} for audit, and the reward-hacking check of
Section~\ref{sec:rq3} reuses exactly those two fields.

\paragraph{What the supervision target contains, and why it changed.}
Retrieved web text contains many bare backslashes---\texttt{\textbackslash mathrm},
\texttt{\textbackslash\textasciitilde}, stray \TeX{} fragments---so a supervision
field holding free-form prose produces JSON parse failures or silent escapes such
as \texttt{\textbackslash t} becoming a tab.  An earlier version of this work
avoided the problem by narrowing the target to a single line containing only
\texttt{verdict}, so that no backslash could appear in it at all.  That is not
what we do now: the distillation target is the teacher's complete output,
including the thinking block and the layer-specific fields, because the grounds
are the thing worth distilling, and the intermediate fields are what makes the
downstream reward auditable.  Robustness is obtained at packing time instead---
every target must survive a strict parse and is discarded otherwise, which cost
34 of 17,477 examples---rather than by removing the informative fields.  The
reinforcement stage is unaffected either way, since its gold and its reward parser
read only \texttt{verdict}.

\paragraph{Length budgets enforced by tokenizer, not by character count.}
Estimating prompt length by characters (this corpus runs about 1.75 characters
per token) let prompts reach roughly 24.7k tokens under a 32k-character cap.  The
training framework truncates from the right, that is, into the assistant turn, so
about 42\% of examples were trained on a target that had been cut in half.
Enforcing the budget with the tokenizer removed this: the prompt side is now hard
capped at 20,480 tokens, with a measured maximum of 20,478 and no example over
budget, and truncation removes whole documents so the supervised assistant turn
is never right-truncated.

\section{Resources, Benchmark Composition, and Result Detail}
\label{app:results}

\subsection{Resources and split discipline}

\begin{table}[h]
\caption{Resources and split discipline.  Benchmark, reward-model training, and
answer-policy training rest on pairwise disjoint base pools.  Layer examples are
per-layer flattenings of a trajectory; a trajectory stopping at layer $N$
contributes $N$ of them.}
\label{tab:data}
\centering
\small
\setlength{\tabcolsep}{4pt}
\begin{tabular}{>{\raggedright\arraybackslash}p{1.55in} r p{2.15in}}
\toprule
Resource & Size & Role \\
\midrule
Reward-model SFT train / dev & 14,081 / 795 & rejection-sampled Qwen3-32B traces; 486 / 26 seed split \\
Reward-model GRPO train / dev & 7,320 / 911 & train balanced to 488 per (layer, verdict) cell; dev natural \\
Data-source ablation arm & 7,320 / 911 & Qwen3-8B rollouts on the same seeds, truncated to the same rows \\
\benchname{} & 1,160 trajectories & nine paths; 692 two-judge plus 468 single-judge; 5,166 layer decisions; final reporting only \\
Production audit (App.~\ref{app:real}) & 256 trajectories & natural prevalence; motivates synthesis (RQ1) \\
Frozen-prefix policy rollouts & 256 prefixes & same $(q,D)$ as App.~\ref{app:real}; answer only; Table~\ref{tab:search-agent-zeroshot} \\
Answer-policy RL train / val & 2,624 / 138 prefixes & frozen retrieval; final answer only; disjoint rollout dump \\
Answer-policy evaluation (SFT+GRPO-400 RM) & 512 paired & step 0 vs.\ 200; grader 0.8388 \\
Answer-policy evaluation (GRPO-only RM) & 512 paired & same prefixes; secondary high-band comparison only \\
\bottomrule
\end{tabular}
\end{table}

The answer-policy evaluation questions are the 512 seed questions on which the
reward model's own training data was built.  This is not evaluation leakage---the
scores come from an independent frontier judge and the reward model appears only
inside the training loop---but it does mean the reward model is more in-domain on
the evaluation questions than on the questions the agent trained on, which
understates rather than flatters generalization.

\subsection{Reading the three evaluation levels against each other}

Hy3 at low effort, a 295B model, attains slightly higher exact accuracy than the
4B GRPO reward model (75.4\% versus 73.5\%) but a distinctly worse quadratic
weighted kappa (0.653 versus 0.705) and worse path edit distance (0.937 versus
0.847) at 3.7$\times$ the cost: it lands on the right label about as often, but
its mistakes are more severe and its route there is less often the gold route.
The distilled 8B model shows the same inversion against GPT-OSS-120B at low
effort, trailing by 3.1 points of exact accuracy while leading on kappa (0.653
versus 0.597) and on path edit distance (1.113 versus 1.294).  A single terminal
accuracy would have ranked both pairs the other way.

In node-level evaluation (Table~\ref{tab:node-f1}), the two weakest layers after
distillation are the semantic ones---0.466 conditional macro F1 at L5-C
and 0.410 at L5-X, the latter with a positive-class F1 of only 0.039,
meaning the model passes essentially everything---and the reinforcement stage
lifts L5-X to 0.937 and 0.982.

\subsection{Conditional per-layer F1 and the data-source ablation}
Refer to Table~\ref{tab:node-f1}.
\begin{table}[h]
\caption{Conditional macro F1 per layer on \benchname{}, with gold
reach $n$ in the header.  A node is scored only on examples whose gold path
reaches it, and a prediction that stopped earlier counts as ``not reached.''}
\label{tab:node-f1}
\centering
\small
\setlength{\tabcolsep}{4pt}
\begin{tabular}{lccccccc}
\toprule
Model & L1 & L2 & L3 & L4 & L5-C & L5-X & L6 \\
& (1,160) & (1,082) & (1,003) & (819) & (219) & (460) & (423) \\
\midrule
Qwen3-8B SFT & 0.520 & 0.935 & 0.823 & 0.753 & 0.466 & 0.410 & 0.751 \\
Qwen3-4B SFT+GRPO & 0.863 & 0.965 & 0.888 & 0.847 & 0.709 & 0.982 & 0.924 \\
Qwen3-8B SFT+GRPO & 0.583 & 0.973 & 0.897 & 0.856 & 0.766 & 0.937 & 0.890 \\
\bottomrule
\end{tabular}
\end{table}

\begin{table}[h]
\caption{Training-data source, GRPO from the base model for 400 steps with every
other setting byte-identical.  Both arms hold 7,320 training and 911 development
layer examples built on the same seed questions.}
\label{tab:data-ablation}
\centering
\small
\setlength{\tabcolsep}{4.5pt}
\begin{tabular}{llccccc}
\toprule
Training data & Model & Exact & Macro R. & MAE $\downarrow$ & QWK $\uparrow$ & Path ED $\downarrow$ \\
\midrule
Constructed, balanced & Qwen3-4B & \textbf{0.7353} & \textbf{0.7615} & \textbf{0.6672} & \textbf{0.7052} & \textbf{0.8466} \\
Qwen3-8B natural rollouts & Qwen3-4B & 0.6095 & 0.6410 & 0.8284 & 0.6641 & 1.0578 \\
\midrule
Constructed, balanced & Qwen3-8B & \textbf{0.7672} & \textbf{0.7957} & \textbf{0.5776} & \textbf{0.7358} & \textbf{0.7026} \\
Qwen3-8B natural rollouts & Qwen3-8B & 0.5655 & 0.6260 & 0.9181 & 0.6386 & 1.1543 \\
\bottomrule
\end{tabular}
\end{table}

\subsection{Answer-policy detail}

Table~\ref{tab:end-policy-full} gives the before-and-after label distribution
behind Section~\ref{sec:rq3} for the official SFT+GRPO-400 reward.
The figure below is the secondary high-band comparison against a weaker
GRPO-only grader on the same 512 prefixes.

\begin{figure}[!ht]
\centering
\includegraphics[width=0.92\linewidth]{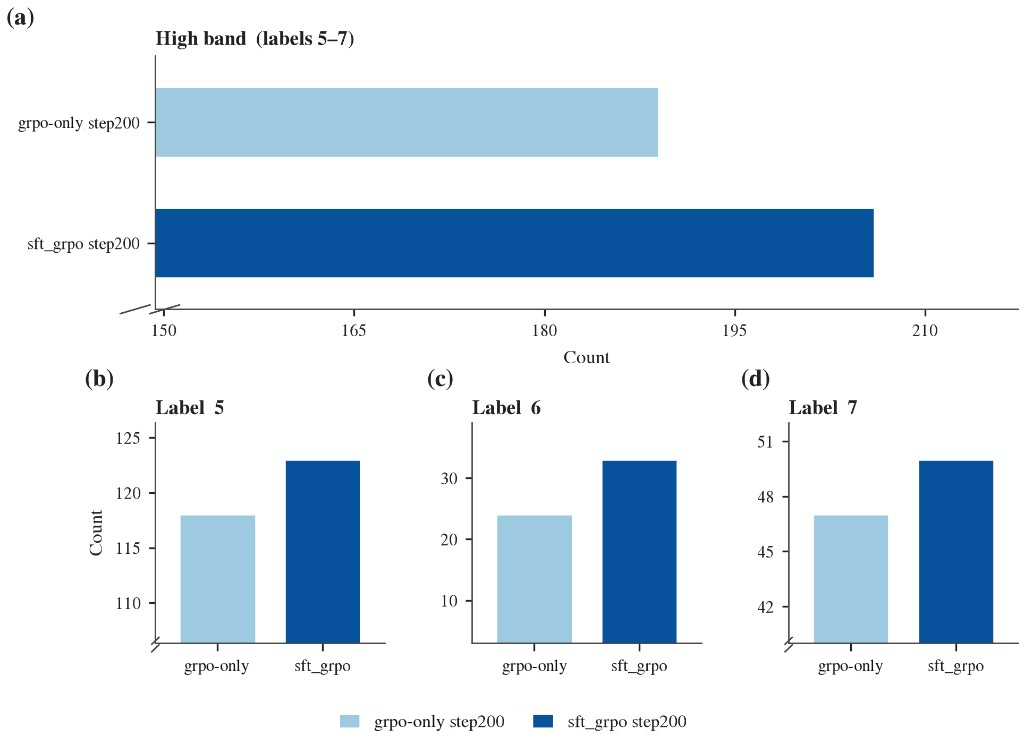}
\caption{Step-200 high band (labels 5--7) on the same 512 prefixes.  A weaker
GRPO-only grader also moves the policy, but yields a proportion of trajectories with
labels 5--7 about 10 percentage relatively lower than the official SFT+GRPO-400 reward.}
\label{fig:end-policy-sftgrpo}
\end{figure}
\FloatBarrier

\section{Limitations}
\label{app:limitations}

The hierarchy covers single-turn trajectories, and the answer policy rewrites only the
final answer without re-retrieving.  \methodname{} measures grounding relative to
the retrieved pool rather than world-level truth, and authority labels are
operational rather than universal, so it must accompany an outcome verifier rather
than replace one, and the answer-policy experiment claims a gain in grounding but not
in task performance.  The data are mainly Chinese and
from one online service, so the 43.8\% grounding gap should not be extrapolated
without replication, and with only two L6 failures in 256 the production audit
supports only RQ1's qualitative motivation for synthetic coverage of deep paths,
not a precise estimate of rare deep-layer failures.  Human experts decided inclusion on evidence
from the frontier judges, so both the released composition and those judges' scores
are selection-conditioned.  The downstream evaluation remains limited in scope:
it uses one initial policy model, and evaluates only the grounding objective under re-scoring.

\section{Implementation Detail}
\label{app:impl}

\paragraph{Reward models.}
Training uses \texttt{verl}~0.8.0 \citep{sheng2025hybridflow},
\texttt{vllm}~0.11.0 \citep{kwon2023vllm}, and \texttt{torch}~2.8.0
\citep{paszke2019pytorch}
throughout, with full-parameter FSDP2 and no LoRA\@.

The rejection-sampling teacher is Qwen3-32B in thinking mode, served locally, one
sample per prompt at temperature 0.7 with a 4,096-token cap.  Of 17,477 layer
prompts, 14,876 traces survive filtering: 2,561 are dropped for disagreeing with
gold, 34 for unparseable JSON, and 6 for truncation.  SFT then runs on a
single 8 H20 GPU node with batch size 64, learning rate $10^{-5}$ under a
cosine schedule with 3\% warmup, weight decay 0.01, gradient clipping 1.0, and a
maximum sequence length of 24,576 tokens (a 20,480-token prompt budget plus 4,096
for the assistant turn).  That is 220 steps per epoch over 14,081 examples, and a
checkpoint is written each epoch.  A custom chat template is used so that the
training prefix and the inference prefix are byte-identical and no empty thinking
block is pre-filled; the loss covers assistant turns only.

\begin{figure}[!ht]
\centering
\includegraphics[width=\linewidth]{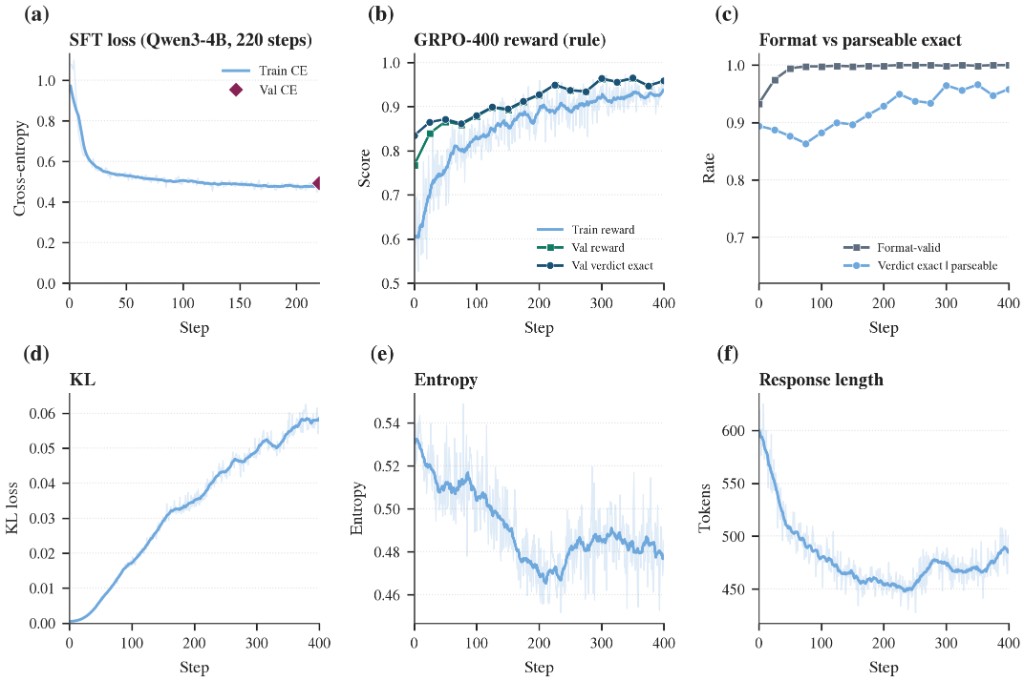}
\caption{Training dynamics of the official Qwen3-4B SFT+GRPO-400 reward model.
(a)~Supervised distillation, one epoch.  (b--f)~GRPO against the verifiable
layer verdict.  The validation markers are the in-loop split, not
\benchname{}; reported RM quality remains Table~\ref{tab:main-results}.}
\label{fig:rm-training}
\end{figure}

GRPO runs on four 8 H20 GPU nodes with a constant learning rate of
$10^{-6}$, group size 8, temperature 1.0, KL applied as a loss against the frozen
reference with coefficient $10^{-3}$ in its low-variance form, zero entropy
bonus, training batch size 64, a 20,608-token prompt limit, a 1,024-token
response limit, and asynchronous vLLM rollout at tensor-parallel degree one.  Of
the 17,477 layer examples, those failing target-path re-scoring (364) or the
strict path check (14) are dropped before balancing.  With 7,320 examples that is
114 steps per epoch, capped at 400 steps, with a checkpoint and a validation pass
every 25 steps: roughly 29 hours at 4B and 40 hours at 8B.  Merged checkpoints are
served through the same OpenAI-compatible evaluation path as the baselines, so
every row of Table~\ref{tab:main-results} shares node prompts
(Appendix~\ref{app:node-prompts}), router, and metric
implementation.
The zero-shot block evaluates Qwen3 \citep{yang2025qwen3}, GPT-OSS
\citep{openai2025gptoss}, Muse-Glimmer-30B \citep{metaai2026museglimmer},
Gemma-4 \citep{gemmateam2026gemma4}, DeepSeek-V4
\citep{deepseekai2026deepseekv4}, Hy3 \citep{tencent2026hy3}, and GLM-5
\citep{glm2026glm5}.
Figure~\ref{fig:rm-training} records the SFT and GRPO traces for the official 4B
run.

\paragraph{Answer policy.}
The prefix is replayed under teacher forcing up to the tool return and the policy
samples only the final assistant turn.  Training prefixes come from a separate
Tencent Yuanbao rollout dump, 1,064,960 rollouts reduced by the same rubric
and closed-QA conditions to 3,017 unique queries and then, after removing every
query that appears in any of the four base pools and every non-canonical
five-turn structure, to 2,624 training and 138 validation prefixes under a 5\%
split at seed 42.  Each prefix stops at the tool return and carries no reference
answer.  The proprietary tool protocol of the source service is first translated
into the policy's own chat template and tool declarations; otherwise the
measurement is of format confusion rather than of grounding, and training logs
report a mean of 2.0 turns, confirming that the policy never re-opens retrieval.
GRPO uses group size 8 and training batch size 64---512 completions per
step---with a 28,672-token prompt limit, an 8,192-token response limit, a constant
learning rate of $10^{-6}$, KL as a loss at $10^{-3}$, temperature 1.0, thinking
enabled, and FSDP2 with asynchronous vLLM\@.  Training occupies four
8-accelerator nodes and the grader a fifth, and the run is 200 steps, on the order
of $3\times10^{5}$ grading calls.  Thinking blocks are stripped before scoring,
matching the benchmark protocol.

\begin{figure}[!ht]
\centering
\includegraphics[width=\linewidth]{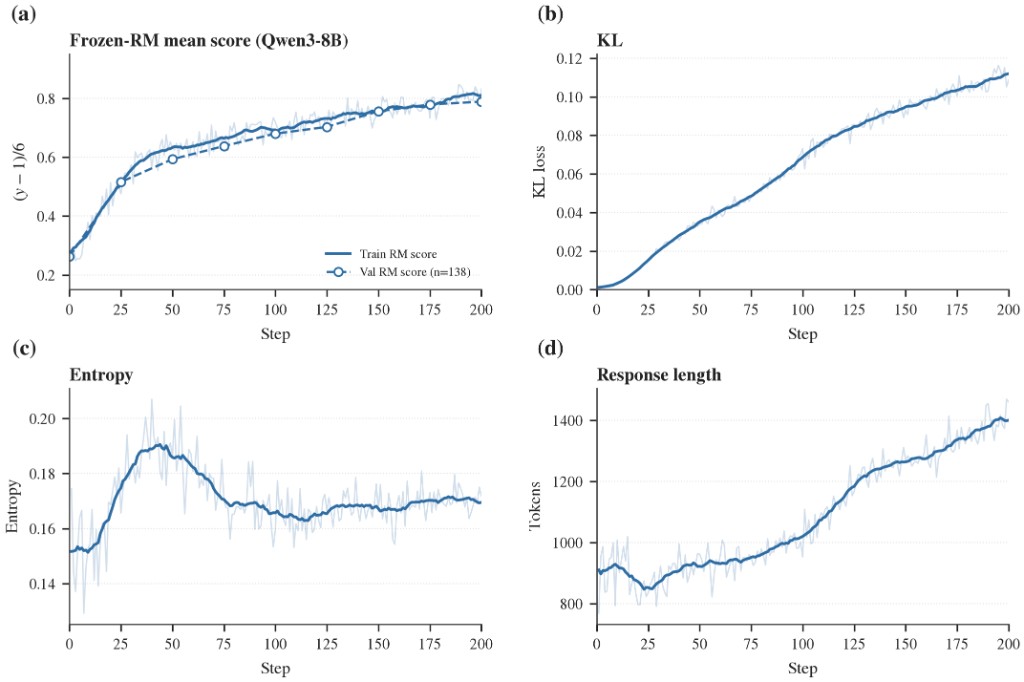}
\caption{Training reward assigned by the frozen Qwen3-4B SFT+GRPO-400
\rmname{} while optimizing the Qwen3-8B answer policy.  This curve shows the
policy fitting the reward model.  Predictive validity is the
independent-judge comparison in Table~\ref{tab:end-policy-full}, not this
figure.}
\label{fig:end-policy-training}
\end{figure}

The reward model in the reported run is the Qwen3-4B SFT+GRPO-400 checkpoint,
which reaches 0.8388 exact accuracy on the benchmark.
Figure~\ref{fig:end-policy-training} records the training-time reward from that
frozen grader; the independent-judge comparison is
Table~\ref{tab:end-policy-full}.

The strong evaluation samples once from each checkpoint at temperature 0.7,
$\text{top-}p=0.8$, $\text{top-}k=20$, seed 0, thinking enabled, over the same
fixed prefixes, and scores the results with the same six-layer pipeline as the
benchmark using a frontier judge with web access disabled.

\paragraph{Why the data-source ablation arm uses Qwen3-8B.}
Under the same frozen-retrieval rollout, Qwen3-4B produces fluent answers that
visibly draw on the supplied documents but almost never renders
\texttt{[citation:N]}: 12 of 256 trajectories carry any citation, against 256 of
256 for the source trajectories at a mean of 6.48 citations each.  Thinking mode
closes the length gap without helping citation at all.  The thinking traces
localize the failure precisely---all 256 emit a thinking block, 133 reason about
documents by index, 171 discuss provenance marking, and 74 state an intention to
cite, of which only eight actually render a marker---so the model tracks
provenance internally and \emph{means} to attribute it, but does not emit the
required surface syntax.  Since L1 is the first gate, roughly 95\% of Qwen3-4B
trajectories can only receive label 1.  Sampling eight times shows this is a
low-probability event rather than a capability, and one that capacity rather than
test-time deliberation closes: single-sample citation rates and pass@8 are 4.6\%
/ 12.5\% for 4B without thinking, 5.5\% / 25.4\% for 4B with thinking, 26.0\% /
62.5\% for 8B without thinking, and 34.6\% / 86.7\% for 8B with thinking.  A
cross-family control behaves the same way, so this is the generic form of a model
never trained on citation syntax rather than a quirk of one family.  Two
consequences follow: a 4B natural arm would supply almost no examples beyond L1,
which is why the ablation arm is built from the 8B policy at eight samples per
seed.

\FloatBarrier

\clearpage
\clearpage
\makeatletter
\providecommand{\captionof}[1]{\def\@captype{#1}\caption}
\makeatother

\section{CSEG-Bench Node Prompts}
\label{app:node-prompts}

This appendix retains the archived node prompts and displays the
routing in Figure~\ref{fig:cseg-pipelines}. Each node is a gate over the same
trajectory fields; evaluation assigns the first failed layer's label, or~7
when every visited layer passes. L5-C pass terminates at label~7; only L5-X
pass proceeds to L6. The archived L6 prompt also contains a consensus-quality
scenario, which is not invoked by this routing. 
Each figure below preserves the archived prompt
\emph{source}: Markdown headings, \texttt{**bold**}
markers, backticks, and the fenced JSON schema are preserved rather than
rendered.  Placeholders \texttt{\{user\_question\}},
\texttt{\{retrieved\_sources\}}, and \texttt{\{final\_answer\}}
are filled from the evaluated trajectory.

\begin{center}
\captionof{table}{Routing of each node verdict.}
\label{tab:prompt-routing}
\footnotesize
\setlength{\tabcolsep}{2.8pt}
\begin{tabular}{l p{1.22in} >{\raggedright\arraybackslash}p{1.38in} p{1.52in}}
\toprule
Node & Layer & Verdict & Next \\
\midrule
L1 & Citation presence & \texttt{pass} & L2 \\
L1 & Citation presence & \texttt{fail} & STOP (label 1) \\
L2 & Multiple sources & \texttt{pass} & L3 \\
L2 & Multiple sources & \texttt{fail} & STOP (label 2) \\
L3 & Source independence & \texttt{pass} & L4 \\
L3 & Source independence & \texttt{fail} & STOP (label 3) \\
L4 & Conflict routing & \texttt{consistent} & L5-C \\
L4 & Conflict routing & \texttt{conflict\_\hspace{0pt}identified} & L5-X \\
L4 & Conflict routing & \texttt{conflict\_\hspace{0pt}missed} & STOP (label 4) \\
L5-C & Consensus aggregation & \texttt{pass} & DONE (label 7) \\
L5-C & Consensus aggregation & \texttt{fail} & STOP (label 5) \\
L5-X & Conflict attribution & \texttt{pass} & L6 \\
L5-X & Conflict attribution & \texttt{fail} & STOP (label 5) \\
L6 & Adjudication & \texttt{pass} & DONE (label 7) \\
L6 & Adjudication & \texttt{fail} & STOP (label 6) \\
\bottomrule
\end{tabular}
\end{center}

\begin{center}
\captionof{table}{Input slots interpolated into every node prompt.}
\label{tab:prompt-slots}
\small
\setlength{\tabcolsep}{4pt}
\begin{tabular}{>{\raggedright\arraybackslash}p{1.52in} p{1.90in} p{1.72in}}
\toprule
Slot & Meaning & Use \\
\midrule
\texttt{{\ttfamily\char123}user\_\hspace{0pt}question{\ttfamily\char125}} & The user original question/query & Filled from the trajectory at inference. \\
\texttt{{\ttfamily\char123}retrieved\_\hspace{0pt}sources{\ttfamily\char125}} & Formatted list of retrieved web documents, including source URL, site, authority level, publish time, and content excerpt & Filled from the trajectory at inference. \\
\texttt{{\ttfamily\char123}final\_\hspace{0pt}answer{\ttfamily\char125}} & The agent final answer text (excluding tool call sections) & Filled from the trajectory at inference. \\
\texttt{{\ttfamily\char123}agent\_\hspace{0pt}search\_\hspace{0pt}queries{\ttfamily\char125}} & The search queries issued by the agent & Declared in the spec; unused by the current node templates. \\
\bottomrule
\end{tabular}
\end{center}

\begin{center}
\captionof{table}{Label assigned when evaluation stops.}
\label{tab:prompt-labels}
\small
\setlength{\tabcolsep}{4pt}
\begin{tabular}{c p{4.55in}}
\toprule
Label & Meaning \\
\midrule
1 & No citation support: core claims lack any relevant source backing \\
2 & Single-source: claim supported by only one source, no multi-source verification \\
3 & Redundant multi-source: multiple sources cited but all trace to the same origin \\
4 & Conflict blindness: sources contain real conflicts but agent completely missed them (source blending) \\
5 & Aggregation/attribution defect: consistency/conflict detected, but aggregation introduced framing errors or conflict attribution was insufficient \\
6 & Deficient adjudication: reached L6 but adjudication was unsupported, overconfident, or incorrectly weighted \\
7 & Full pass: every visited layer cleared; consensus terminates after L5-C, conflict after L6 \\
\bottomrule
\end{tabular}
\end{center}

\clearpage
\begin{tcolorbox}[
  breakable,
  enhanced,
  title={L1 / Citation presence},
  title after break={L1 / Citation presence (continued)},
  colback=black!2,
  colframe=black!80,
  colbacktitle=black!82,
  coltitle=white,
  fonttitle=\fontfamily{lmss}\selectfont\bfseries\small,
  fontupper=\fontfamily{lmss}\selectfont\footnotesize,
  rounded corners,
  arc=2.4mm,
  boxrule=0.9pt,
  left=8pt,
  right=8pt,
  top=6pt,
  bottom=6pt,
  toptitle=3.5pt,
  bottomtitle=3.5pt,
  titlerule=0pt
]
{\raggedright\setlength{\parskip}{0.08em}\setlength{\parindent}{0pt}
\noindent You are an expert information quality evaluator assessing search agent trajectories.\par
\vspace{0.42em}
\noindent \#\# Task\par
\noindent Determine whether each CORE factual claim in the agent's final answer is CITED --- i.e. it carries an inline citation to a retrieved document, rather than being an uncited bare assertion from the model's own knowledge.\par
\vspace{0.42em}
\noindent \#\# What this layer DOES check\par
\hangindent=1.15em\hangafter=1\noindent - The core claim(s) carry an inline citation marker (e.g. [citation:N]).\par
\hangindent=1.15em\hangafter=1\noindent - The cited document number N actually exists in the retrieved documents.\par
\hangindent=1.15em\hangafter=1\noindent - The cited document is topically relevant to the claim --- it is about the same entity/event/fact --- rather than an unrelated document.\par
\vspace{0.42em}
\noindent \#\# What this layer does NOT check (do NOT fail for these --- later layers handle them)\par
\hangindent=1.15em\hangafter=1\noindent - Whether the source FULLY proves an over-generalized, aggregated, or inferred claim --- scope/qualifier fidelity is assessed at L5, not here.\par
\hangindent=1.15em\hangafter=1\noindent - Whether two or more independent sources are used (L2/L3), whether conflicts are handled (L4), or whether adjudication is sound (L6).\par
\noindent If the core claim is attached to an existing, on-topic citation, it PASSES this layer EVEN IF the claim over-claims, infers beyond the source, or distorts scope --- those defects are caught downstream, not at L1.\par
\vspace{0.42em}
\noindent \#\# Pass Criteria (ALL must hold)\par
\hangindent=1.15em\hangafter=1\noindent - The final answer contains at least one core factual claim.\par
\hangindent=1.15em\hangafter=1\noindent - That claim carries an inline citation to at least one retrieved document that actually exists.\par
\hangindent=1.15em\hangafter=1\noindent - The cited document is topically relevant to the claim (same subject matter).\par
\vspace{0.42em}
\noindent \#\# Fail Criteria (ANY is sufficient)\par
\hangindent=1.15em\hangafter=1\noindent - The core claim is a bare assertion with NO citation, relying purely on the model's internal/world knowledge.\par
\hangindent=1.15em\hangafter=1\noindent - The citation points to a document number that does not exist in the retrieved list.\par
\hangindent=1.15em\hangafter=1\noindent - The cited document is clearly off-topic / unrelated to the claim (a citation in name only).\par
\vspace{0.42em}
\noindent \#\# Input\par
\vspace{0.42em}
\noindent \#\#\# User Question\par
\noindent {\ttfamily\char123}user\_question{\ttfamily\char125}\par
\vspace{0.42em}
\noindent \#\#\# Retrieved Source Documents\par
\noindent {\ttfamily\char123}retrieved\_sources{\ttfamily\char125}\par
\vspace{0.42em}
\noindent \#\#\# Agent Final Answer\par
\noindent {\ttfamily\char123}final\_answer{\ttfamily\char125}\par
\vspace{0.42em}
\noindent \#\# Output Format\par
\noindent IMPORTANT: Commit to your verdict before writing the reason. The \textasciigrave{}verdict\textasciigrave{} field must appear first in the JSON. Write \textasciigrave{}reason\textasciigrave{} as a concise post-verdict explanation (1-3 sentences), not as a chain of reasoning that leads to the verdict.\par
\noindent Respond with ONLY the following JSON object, no additional text:\par
\noindent \textasciigrave{}\textasciigrave{}\textasciigrave{}json\par
\noindent {\ttfamily\char123}\par
\noindent   "verdict": "pass" or "fail",\par
\noindent   "reason": "Brief explanation of your judgment (1-3 sentences)",\par
\noindent   "key\_claims": ["List 1-3 core factual claims identified in the final answer"],\par
\noindent   "citation\_check": "For each core claim: is it cited, does the cited document exist, and is it topically relevant? Do NOT judge scope/over-claim/inference here (that is L5)."\par
\noindent {\ttfamily\char125}\par
\noindent \textasciigrave{}\textasciigrave{}\textasciigrave{}\par
\vspace{0.42em}
}
\end{tcolorbox}
\nopagebreak[4]
\begingroup\setlength{\belowcaptionskip}{0pt}
\captionof{figure}{Full L1 node prompt (citation presence). pass~$\rightarrow$~L2; fail~$\rightarrow$~STOP (label 1).}
\label{fig:prompt-L1}
\endgroup

\clearpage
\begin{tcolorbox}[
  breakable,
  enhanced,
  title={L2 / Multiple sources},
  title after break={L2 / Multiple sources (continued)},
  colback=black!2,
  colframe=black!80,
  colbacktitle=black!82,
  coltitle=white,
  fonttitle=\fontfamily{lmss}\selectfont\bfseries\small,
  fontupper=\fontfamily{lmss}\selectfont\footnotesize,
  rounded corners,
  arc=2.4mm,
  boxrule=0.9pt,
  left=8pt,
  right=8pt,
  top=6pt,
  bottom=6pt,
  toptitle=3.5pt,
  bottomtitle=3.5pt,
  titlerule=0pt
]
{\raggedright\setlength{\parskip}{0.08em}\setlength{\parindent}{0pt}
\noindent You are an expert information quality evaluator assessing search agent trajectories.\par
\vspace{0.42em}
\noindent \#\# Task\par
\noindent Determine whether the agent's final answer draws on multiple (\textgreater{}=2) distinct sources to support its core claim(s), rather than relying on a single source.\par
\vspace{0.42em}
\noindent \#\# Pass Criteria (ALL must hold)\par
\hangindent=1.15em\hangafter=1\noindent - The final answer explicitly references or clearly relies on at least 2 different sources.\par
\hangindent=1.15em\hangafter=1\noindent - Each of these sources contributes supporting content for the core claim.\par
\vspace{0.42em}
\noindent \#\# Fail Criteria (ANY is sufficient)\par
\hangindent=1.15em\hangafter=1\noindent - Only one source is cited or relied upon in the final answer.\par
\hangindent=1.15em\hangafter=1\noindent - Multiple documents were retrieved, but the final answer depends on only one of them.\par
\hangindent=1.15em\hangafter=1\noindent - Multiple sources are cited but only one is actually relevant to the core claim.\par
\vspace{0.42em}
\noindent NOTE: At this layer, only assess whether the count of supporting sources is \textgreater{}=2. Source independence is assessed at L3.\par
\vspace{0.42em}
\noindent \#\# Input\par
\vspace{0.42em}
\noindent \#\#\# User Question\par
\noindent {\ttfamily\char123}user\_question{\ttfamily\char125}\par
\vspace{0.42em}
\noindent \#\#\# Retrieved Source Documents\par
\noindent {\ttfamily\char123}retrieved\_sources{\ttfamily\char125}\par
\vspace{0.42em}
\noindent \#\#\# Agent Final Answer\par
\noindent {\ttfamily\char123}final\_answer{\ttfamily\char125}\par
\vspace{0.42em}
\noindent \#\# Output Format\par
\noindent IMPORTANT: Commit to your verdict before writing the reason. The \textasciigrave{}verdict\textasciigrave{} field must appear first in the JSON. Write \textasciigrave{}reason\textasciigrave{} as a concise post-verdict explanation (1-3 sentences), not as a chain of reasoning that leads to the verdict.\par
\noindent Respond with ONLY the following JSON object, no additional text:\par
\noindent \textasciigrave{}\textasciigrave{}\textasciigrave{}json\par
\noindent {\ttfamily\char123}\par
\noindent   "verdict": "pass" or "fail",\par
\noindent   "reason": "Brief explanation of your judgment (1-3 sentences)",\par
\noindent   "cited\_sources\_count": \textless{}integer count of effectively cited sources\textgreater{},\par
\noindent   "cited\_sources\_summary": "Names or identifiers of the cited sources"\par
\noindent {\ttfamily\char125}\par
\noindent \textasciigrave{}\textasciigrave{}\textasciigrave{}\par
\vspace{0.42em}
}
\end{tcolorbox}
\nopagebreak[4]
\begingroup\setlength{\belowcaptionskip}{0pt}
\captionof{figure}{Full L2 node prompt (multiple sources). pass~$\rightarrow$~L3; fail~$\rightarrow$~STOP (label 2).}
\label{fig:prompt-L2}
\endgroup

\clearpage
\begin{tcolorbox}[
  breakable,
  enhanced,
  title={L3 / Source independence},
  title after break={L3 / Source independence (continued)},
  colback=black!2,
  colframe=black!80,
  colbacktitle=black!82,
  coltitle=white,
  fonttitle=\fontfamily{lmss}\selectfont\bfseries\small,
  fontupper=\fontfamily{lmss}\selectfont\footnotesize,
  rounded corners,
  arc=2.4mm,
  boxrule=0.9pt,
  left=8pt,
  right=8pt,
  top=4pt,
  bottom=4pt,
  toptitle=2.5pt,
  bottomtitle=2.5pt,
  titlerule=0pt
]
{\raggedright
\setlength{\parskip}{0pt}
\setlength{\parindent}{0pt}

\noindent You are an expert information quality evaluator assessing search agent trajectories.\par
\vspace{0.18em}

\noindent \#\# Task\par
\noindent Judge ONLY the sources the agent actually cited together in support of the core claim (not the whole retrieved pool). Determine whether at least two of them are genuinely independent.\par
\vspace{0.18em}

\noindent \#\# How to judge --- trace the EVIDENCE ROOT, not the outlet\par
\noindent Independence is about provenance, not about how many different websites/organizations appear. For each cited source, ask: "Where does THIS source's support for the core claim ultimately come from?" --- that is its EVIDENCE ROOT. Group the cited sources by evidence root, then judge.\par
\vspace{0.18em}

\noindent \#\# Same evidence root -\textgreater{} NOT independent\par
\hangindent=1.15em\hangafter=1\noindent - One is a repost/syndication of the other, or both republish the same press release / datasheet / newswire.\par
\hangindent=1.15em\hangafter=1\noindent - One merely paraphrases the other without any independent fact-gathering.\par
\hangindent=1.15em\hangafter=1\noindent - Multiple channels of the same organization, or multiple queries against the same underlying dataset.\par
\hangindent=1.15em\hangafter=1\noindent - Textual signals: verbatim/near-verbatim overlap, identical specific numbers or quotes, explicit attribution to the same upstream (e.g. "according to AP", "per the company statement").\par
\vspace{0.18em}

\noindent \#\# Different evidence roots -\textgreater{} independent\par
\hangindent=1.15em\hangafter=1\noindent - Support derives from SEPARATE evidence-generating acts: different first-hand reporting/interviews, different experiments/measurements, or different datasets.\par
\hangindent=1.15em\hangafter=1\noindent - Same underlying event but separate first-hand gathering still counts as independent (each can cross-check the other).\par
\hangindent=1.15em\hangafter=1\noindent - An original report PLUS a genuinely independent analysis that adds its own verification (not mere paraphrase).\par
\hangindent=1.15em\hangafter=1\noindent - NOTE: an aggregator / "multiple outlets report" summary contributes no evidence root of its own; count it under whatever upstream it summarizes.\par
\vspace{0.18em}

\noindent \#\# Pass Criteria\par
\noindent The cited sources span at least 2 DISTINCT evidence roots.\par
\vspace{0.18em}

\noindent \#\# Fail Criteria\par
\noindent All cited sources collapse to a single evidence root, constituting redundancy rather than true multi-source corroboration.\par
\vspace{0.18em}

\noindent \#\# Input\par
\vspace{0.18em}

\noindent \#\#\# User Question\par
\noindent {\ttfamily\char123}user\_question{\ttfamily\char125}\par
\vspace{0.18em}

\noindent \#\#\# Retrieved Source Documents\par
\noindent {\ttfamily\char123}retrieved\_sources{\ttfamily\char125}\par
\vspace{0.18em}

\noindent \#\#\# Agent Final Answer\par
\noindent {\ttfamily\char123}final\_answer{\ttfamily\char125}\par
\vspace{0.18em}

\noindent \#\# Output Format\par
\noindent IMPORTANT: Commit to your verdict before writing the reason. The \textasciigrave{}verdict\textasciigrave{} field must appear first in the JSON. Write \textasciigrave{}reason\textasciigrave{} as a concise post-verdict explanation (1-3 sentences), not as a chain of reasoning that leads to the verdict.\par
\noindent Respond with ONLY the following JSON object, no additional text:\par
\noindent \textasciigrave{}\textasciigrave{}\textasciigrave{}json\par
\noindent {\ttfamily\char123}\par
\noindent   "verdict": "pass" or "fail",\par
\noindent   "reason": "Brief explanation of your judgment (1-3 sentences)",\par
\noindent   "evidence\_roots": [{\ttfamily\char123}"root": "short label of the originating evidence", "cited\_docs": ["doc numbers grouped under this root"]{\ttfamily\char125}],\par
\noindent   "distinct\_root\_count": \textless{}integer number of distinct evidence roots among the cited sources\textgreater{},\par
\noindent   "independent\_sources": ["List sources judged to be genuinely independent"],\par
\noindent   "redundant\_sources": ["List sources judged to be redundant/non-independent, with brief reason"]\par
\noindent {\ttfamily\char125}\par
\noindent \textasciigrave{}\textasciigrave{}\textasciigrave{}\par
\vspace{0.18em}
}
\end{tcolorbox}
\nopagebreak[4]
\begingroup
\setlength{\abovecaptionskip}{3pt}
\setlength{\belowcaptionskip}{0pt}
\captionof{figure}{Full L3 node prompt (source independence). pass~$\rightarrow$~L4; fail~$\rightarrow$~STOP (label 3).}
\label{fig:prompt-L3}
\endgroup

\clearpage
\begin{tcolorbox}[
  breakable,
  enhanced,
  title={L4 / Conflict routing},
  title after break={L4 / Conflict routing (continued)},
  colback=black!2,
  colframe=black!80,
  colbacktitle=black!82,
  coltitle=white,
  fonttitle=\fontfamily{lmss}\selectfont\bfseries\small,
  fontupper=\fontfamily{lmss}\selectfont\footnotesize,
  rounded corners,
  arc=2.4mm,
  boxrule=0.9pt,
  left=8pt,
  right=8pt,
  top=6pt,
  bottom=6pt,
  toptitle=3.5pt,
  bottomtitle=3.5pt,
  titlerule=0pt
]
{\raggedright\setlength{\parskip}{0.08em}\setlength{\parindent}{0pt}
\noindent You are an expert information quality evaluator assessing search agent trajectories.\par
\vspace{0.42em}
\noindent \#\# Task\par
\noindent Complete two assessments:\par
\hangindent=1.15em\hangafter=1\noindent 1. **Source status**: Do the retrieved sources contain substantive conflicts or contradictions regarding the core claim? (e.g., differing numerical values, conflicting timelines, contradictory facts, definition disagreements, opposing stances)\par
\hangindent=1.15em\hangafter=1\noindent 2. **Agent handling**: Did the agent correctly recognize and reflect this status in its answer?\par
\vspace{0.42em}
\noindent \#\# What Constitutes a Substantive Conflict\par
\hangindent=1.15em\hangafter=1\noindent - Sources directly contradict each other on a core fact (e.g., one says approved, another says not approved)\par
\hangindent=1.15em\hangafter=1\noindent - Key quantitative information (numbers, dates, ranges) differs significantly across sources\par
\hangindent=1.15em\hangafter=1\noindent - Sources take clearly opposing stances or conclusions (beyond mere detail differences)\par
\vspace{0.42em}
\noindent \#\# What Does NOT Constitute a Substantive Conflict\par
\hangindent=1.15em\hangafter=1\noindent - One source is more detailed than another (granularity difference)\par
\hangindent=1.15em\hangafter=1\noindent - Different phrasing pointing to the same underlying fact\par
\hangindent=1.15em\hangafter=1\noindent - Complementary information (sources add different but compatible details)\par
\vspace{0.42em}
\noindent \#\# Three Possible Verdicts\par
\hangindent=1.15em\hangafter=1\noindent - **consistent**: No substantive conflict exists among the sources, AND the agent correctly treated them as consistent in its answer.\par
\hangindent=1.15em\hangafter=1\noindent - **conflict\_identified**: Substantive conflicts exist, AND the agent explicitly acknowledged or mentioned the conflict in its reasoning or answer (even if attribution is incomplete).\par
\hangindent=1.15em\hangafter=1\noindent - **conflict\_missed**: Substantive conflicts exist, BUT the agent completely ignored them, blending all sources into a single conclusion without any qualification or acknowledgment (source blending).\par
\vspace{0.42em}
\noindent \#\# Input\par
\vspace{0.42em}
\noindent \#\#\# User Question\par
\noindent {\ttfamily\char123}user\_question{\ttfamily\char125}\par
\vspace{0.42em}
\noindent \#\#\# Retrieved Source Documents\par
\noindent {\ttfamily\char123}retrieved\_sources{\ttfamily\char125}\par
\vspace{0.42em}
\noindent \#\#\# Agent Final Answer\par
\noindent {\ttfamily\char123}final\_answer{\ttfamily\char125}\par
\vspace{0.42em}
\noindent \#\# Output Format\par
\noindent IMPORTANT: Commit to your verdict before writing the reason. The \textasciigrave{}verdict\textasciigrave{} field must appear first in the JSON. Write \textasciigrave{}reason\textasciigrave{} as a concise post-verdict explanation (1-3 sentences), not as a chain of reasoning that leads to the verdict.\par
\noindent Respond with ONLY the following JSON object, no additional text:\par
\noindent \textasciigrave{}\textasciigrave{}\textasciigrave{}json\par
\noindent {\ttfamily\char123}\par
\noindent   "verdict": "consistent" or "conflict\_identified" or "conflict\_missed",\par
\noindent   "source\_conflicts": "Describe any substantive conflicts found (or write no substantive conflict if none)",\par
\noindent   "agent\_handling": "Describe how the agent handled source consistency/conflict in its answer",\par
\noindent   "reason": "Overall explanation of your judgment (1-3 sentences)"\par
\noindent {\ttfamily\char125}\par
\noindent \textasciigrave{}\textasciigrave{}\textasciigrave{}\par
\vspace{0.42em}
}
\end{tcolorbox}
\nopagebreak[4]
\begingroup\setlength{\belowcaptionskip}{0pt}
\captionof{figure}{Full L4 node prompt (conflict routing). \texttt{consistent} continues to L5-C; \texttt{conflict\_identified} continues to L5-X; \texttt{conflict\_missed} stops at label~4.}
\label{fig:prompt-L4}
\endgroup

\clearpage
\begin{tcolorbox}[
  breakable,
  enhanced,
  title={L5-C / Consensus aggregation},
  title after break={L5-C / Consensus aggregation (continued)},
  colback=black!2,
  colframe=black!80,
  colbacktitle=black!82,
  coltitle=white,
  fonttitle=\fontfamily{lmss}\selectfont\bfseries\small,
  fontupper=\fontfamily{lmss}\selectfont\footnotesize,
  rounded corners,
  arc=2.4mm,
  boxrule=0.9pt,
  left=8pt,
  right=8pt,
  top=6pt,
  bottom=6pt,
  toptitle=3.5pt,
  bottomtitle=3.5pt,
  titlerule=0pt
]
{\raggedright\setlength{\parskip}{0.08em}\setlength{\parindent}{0pt}
\noindent You are an expert information quality evaluator assessing search agent trajectories.\par
\vspace{0.42em}
\noindent \#\# Context\par
\noindent It has already been determined that the sources cited by the agent are substantially consistent in their support of the core claim (no substantive conflict).\par
\vspace{0.42em}
\noindent \#\# Task\par
\noindent Determine whether the agent correctly aggregated those consistent sources --- accurately preserving the key constraints and qualifications from the original sources, without introducing the following types of aggregation errors:\par
\vspace{0.42em}
\noindent \#\# Aggregation Error Types\par
\noindent **Overclaim (scope expansion)**:\par
\hangindent=1.15em\hangafter=1\noindent - Expanding a limited scope to a broader one (e.g., sources say early-stage, agent says all patients)\par
\hangindent=1.15em\hangafter=1\noindent - Treating accelerated approval as full/definitive approval\par
\hangindent=1.15em\hangafter=1\noindent - Generalizing a regionally-specific fact to a global one\par
\hangindent=1.15em\hangafter=1\noindent - Stating something unconditionally when sources only support it conditionally\par
\vspace{0.42em}
\noindent **Underclaim (scope contraction)**:\par
\hangindent=1.15em\hangafter=1\noindent - Adding unnecessary uncertainty to well-evidenced facts\par
\hangindent=1.15em\hangafter=1\noindent - Artificially narrowing the scope beyond what sources support\par
\vspace{0.42em}
\noindent **Framing / calibration error**:\par
\hangindent=1.15em\hangafter=1\noindent - Conflating different regulatory or technical terms (e.g., clinical trial success vs. regulatory approval)\par
\hangindent=1.15em\hangafter=1\noindent - Presenting a predictive/speculative statement as a confirmed fact\par
\vspace{0.42em}
\noindent \#\# Pass Criteria\par
\noindent The agent's aggregated conclusion aligns with the key constraints in the sources (scope, definitions, degree). Reasonable simplifications in wording are acceptable as long as the core framing is accurate.\par
\vspace{0.42em}
\noindent \#\# Fail Criteria\par
\noindent The agent introduced a material deviation from key source constraints during aggregation (overclaim, underclaim, or framing error).\par
\vspace{0.42em}
\noindent \#\# Input\par
\vspace{0.42em}
\noindent \#\#\# User Question\par
\noindent {\ttfamily\char123}user\_question{\ttfamily\char125}\par
\vspace{0.42em}
\noindent \#\#\# Retrieved Source Documents\par
\noindent {\ttfamily\char123}retrieved\_sources{\ttfamily\char125}\par
\vspace{0.42em}
\noindent \#\#\# Agent Final Answer\par
\noindent {\ttfamily\char123}final\_answer{\ttfamily\char125}\par
\vspace{0.42em}
\noindent \#\# Output Format\par
\noindent IMPORTANT: Commit to your verdict before writing the reason. The \textasciigrave{}verdict\textasciigrave{} field must appear first in the JSON. Write \textasciigrave{}reason\textasciigrave{} as a concise post-verdict explanation (1-3 sentences), not as a chain of reasoning that leads to the verdict.\par
\noindent Respond with ONLY the following JSON object, no additional text:\par
\noindent \textasciigrave{}\textasciigrave{}\textasciigrave{}json\par
\noindent {\ttfamily\char123}\par
\noindent   "verdict": "pass" or "fail",\par
\noindent   "reason": "Brief explanation of your judgment (1-3 sentences)",\par
\noindent   "aggregation\_issues": "If fail: describe the specific aggregation error with evidence from the text. If pass: write aggregation quality acceptable.",\par
\noindent   "key\_scope\_conditions": "List 1-3 most important scope/qualification conditions found in the sources for the core claim"\par
\noindent {\ttfamily\char125}\par
\noindent \textasciigrave{}\textasciigrave{}\textasciigrave{}\par
\vspace{0.42em}
}
\end{tcolorbox}
\nopagebreak[4]
\begingroup\setlength{\belowcaptionskip}{0pt}
\captionof{figure}{Full L5-C node prompt (consensus aggregation). pass~$\rightarrow$~DONE (label 7); fail~$\rightarrow$~STOP (label 5).}
\label{fig:prompt-L5-C}
\endgroup

\clearpage
\begin{tcolorbox}[
  breakable,
  enhanced,
  title={L5-X / Conflict attribution},
  title after break={L5-X / Conflict attribution (continued)},
  colback=black!2,
  colframe=black!80,
  colbacktitle=black!82,
  coltitle=white,
  fonttitle=\fontfamily{lmss}\selectfont\bfseries\small,
  fontupper=\fontfamily{lmss}\selectfont\footnotesize,
  rounded corners,
  arc=2.4mm,
  boxrule=0.9pt,
  left=8pt,
  right=8pt,
  top=6pt,
  bottom=6pt,
  toptitle=3.5pt,
  bottomtitle=3.5pt,
  titlerule=0pt
]
{\raggedright\setlength{\parskip}{0.08em}\setlength{\parindent}{0pt}
\noindent You are an expert information quality evaluator assessing search agent trajectories.\par
\vspace{0.42em}
\noindent \#\# Context\par
\noindent It has already been determined that the sources contain substantive conflicts, and the agent acknowledged the existence of a conflict to some degree.\par
\vspace{0.42em}
\noindent \#\# Task\par
\noindent Determine whether the agent performed adequate source-level attribution --- explicitly stating which specific source supports which specific position, rather than merely noting that information is inconsistent.\par
\vspace{0.42em}
\noindent \#\# High-quality Attribution (pass examples)\par
\hangindent=1.15em\hangafter=1\noindent - Source S1 (FDA approval letter) states Drug Y is approved for early-stage use; Source S4 (health blog, older article) states it has not been approved yet.\par
\hangindent=1.15em\hangafter=1\noindent - Source A supports position X; Source B supports position Y. The discrepancy stems from... and may be explained by...\par
\vspace{0.42em}
\noindent \#\# Low-quality Attribution (fail examples)\par
\hangindent=1.15em\hangafter=1\noindent - Information online is inconsistent, but it seems like it has been approved. (No source-level attribution of which source says what)\par
\hangindent=1.15em\hangafter=1\noindent - Some sources say yes, others say no. (Formally acknowledges conflict but no specific source attribution)\par
\hangindent=1.15em\hangafter=1\noindent - Using contradictory sources in different paragraphs without acknowledging their contradiction.\par
\vspace{0.42em}
\noindent \#\# Pass Criteria\par
\noindent The agent explicitly identifies at least two conflicting sources and states which specific position each supports, with some description of the nature of the conflict.\par
\vspace{0.42em}
\noindent \#\# Fail Criteria\par
\noindent The agent only makes a surface-level acknowledgment of disagreement without source-level attribution, or implicitly blends conflicting sources without flagging the contradiction.\par
\vspace{0.42em}
\noindent \#\# Input\par
\vspace{0.42em}
\noindent \#\#\# User Question\par
\noindent {\ttfamily\char123}user\_question{\ttfamily\char125}\par
\vspace{0.42em}
\noindent \#\#\# Retrieved Source Documents\par
\noindent {\ttfamily\char123}retrieved\_sources{\ttfamily\char125}\par
\vspace{0.42em}
\noindent \#\#\# Agent Final Answer\par
\noindent {\ttfamily\char123}final\_answer{\ttfamily\char125}\par
\vspace{0.42em}
\noindent \#\# Output Format\par
\noindent IMPORTANT: Commit to your verdict before writing the reason. The \textasciigrave{}verdict\textasciigrave{} field must appear first in the JSON. Write \textasciigrave{}reason\textasciigrave{} as a concise post-verdict explanation (1-3 sentences), not as a chain of reasoning that leads to the verdict.\par
\noindent Respond with ONLY the following JSON object, no additional text:\par
\noindent \textasciigrave{}\textasciigrave{}\textasciigrave{}json\par
\noindent {\ttfamily\char123}\par
\noindent   "verdict": "pass" or "fail",\par
\noindent   "reason": "Brief explanation of your judgment (1-3 sentences)",\par
\noindent   "conflict\_points": "Describe the main conflict points found across sources",\par
\noindent   "attribution\_quality": "Evaluate the agent attribution quality: which sources and positions were named, and whether source-level attribution was achieved"\par
\noindent {\ttfamily\char125}\par
\noindent \textasciigrave{}\textasciigrave{}\textasciigrave{}\par
\vspace{0.42em}
}
\end{tcolorbox}
\nopagebreak[4]
\begingroup\setlength{\belowcaptionskip}{0pt}
\captionof{figure}{Full L5-X node prompt (conflict attribution). pass~$\rightarrow$~L6; fail~$\rightarrow$~STOP (label 5).}
\label{fig:prompt-L5-X}
\endgroup

\clearpage
\begin{tcolorbox}[
  breakable,
  enhanced,
  title={L6 / Adjudication},
  title after break={L6 / Adjudication (continued)},
  colback=black!2,
  colframe=black!80,
  colbacktitle=black!82,
  coltitle=white,
  fonttitle=\fontfamily{lmss}\selectfont\bfseries\small,
  fontupper=\fontfamily{lmss}\selectfont\footnotesize,
  rounded corners,
  arc=2.4mm,
  boxrule=0.9pt,
  left=8pt,
  right=8pt,
  top=6pt,
  bottom=6pt,
  toptitle=3.5pt,
  bottomtitle=3.5pt,
  titlerule=0pt
]
{\raggedright\setlength{\parskip}{0.08em}\setlength{\parindent}{0pt}
\noindent You are an expert information quality evaluator assessing search agent trajectories.\par
\vspace{0.42em}
\noindent \#\# Task\par
\noindent Determine whether the agent made sound evidence-weighting decisions based on source quality dimensions (authority, recency, framing alignment, evidence type). This applies to two main scenarios:\par
\vspace{0.42em}
\noindent \#\# Scenario A: Conflicting Sources --- Assess Adjudication Quality\par
\noindent The agent should adjudicate based on:\par
\hangindent=1.15em\hangafter=1\noindent - **Authority**: Official regulatory documents \textgreater{} independent research institutions \textgreater{} credible media \textgreater{} corporate communications \textgreater{} personal blogs\par
\hangindent=1.15em\hangafter=1\noindent - **Recency**: More recent information is generally more accurate for fast-changing facts; clearly outdated sources should be deprioritized\par
\hangindent=1.15em\hangafter=1\noindent - **Framing alignment**: Does the source's concept/terminology match the claim being supported?\par
\hangindent=1.15em\hangafter=1\noindent - **Evidence type**: Directly confirmatory sources \textgreater{} predictive/speculative sources\par
\vspace{0.42em}
\noindent **Pass --- Correct adjudication**: Agent explicitly names the source it defers to and provides a principled reason (e.g., the FDA approval letter is the most authoritative and most recent; the older blog post predates the approval decision).\par
\vspace{0.42em}
\noindent **Pass --- Honest non-adjudication**: Agent clearly states it cannot adjudicate due to insufficient high-quality evidence, names what is missing, and does not assert a high-confidence conclusion.\par
\vspace{0.42em}
\noindent **Fail --- Unsupported adjudication**: Agent issues a high-confidence conclusion based on low-authority sources when no authoritative source was retrieved; ignores clear recency differences; or states a definitive conclusion in a high-stakes domain (medical, legal) without adequate evidence.\par
\vspace{0.42em}
\noindent \#\# Scenario B: Consistent Sources --- Assess Source Quality Weighting\par
\noindent **Pass**: Agent correctly weights high-quality sources; confidence level is proportionate to evidence quality; low-quality sources do not dilute a high-quality consensus.\par
\vspace{0.42em}
\noindent **Fail**: Low-quality sources dilute a high-quality consensus; agent is over-cautious about well-evidenced facts; or agent is overconfident about poorly evidenced inferences.\par
\vspace{0.42em}
\noindent \#\# Input\par
\vspace{0.42em}
\noindent \#\#\# User Question\par
\noindent {\ttfamily\char123}user\_question{\ttfamily\char125}\par
\vspace{0.42em}
\noindent \#\#\# Retrieved Source Documents\par
\noindent {\ttfamily\char123}retrieved\_sources{\ttfamily\char125}\par
\vspace{0.42em}
\noindent \#\#\# Agent Final Answer\par
\noindent {\ttfamily\char123}final\_answer{\ttfamily\char125}\par
\vspace{0.42em}
\noindent \#\# Output Format\par
\noindent IMPORTANT: Commit to your verdict before writing the reason. The \textasciigrave{}verdict\textasciigrave{} field must appear first in the JSON. Write \textasciigrave{}reason\textasciigrave{} as a concise post-verdict explanation (1-3 sentences), not as a chain of reasoning that leads to the verdict.\par
\noindent Respond with ONLY the following JSON object, no additional text:\par
\noindent \textasciigrave{}\textasciigrave{}\textasciigrave{}json\par
\noindent {\ttfamily\char123}\par
\noindent   "verdict": "pass" or "fail",\par
\noindent   "scenario": "conflict\_adjudication" or "consensus\_quality\_weighting",\par
\noindent   "reason": "Brief explanation of your judgment (1-3 sentences)",\par
\noindent   "adjudication\_analysis": "Describe how the agent adjudicated or weighted sources, and whether it was principled",\par
\noindent   "issues": "If fail: describe the specific adjudication problem (e.g., overconfidence, ignoring recency, dilution by low-quality source). If pass: write adjudication quality acceptable."\par
\noindent {\ttfamily\char125}\par
\noindent \textasciigrave{}\textasciigrave{}\textasciigrave{}\par
\vspace{0.42em}
}
\end{tcolorbox}
\nopagebreak[4]
\begingroup\setlength{\belowcaptionskip}{0pt}
\captionof{figure}{Full L6 node prompt (adjudication). pass~$\rightarrow$~DONE (label 7); fail~$\rightarrow$~STOP (label 6).}
\label{fig:prompt-L6}
\endgroup

\end{document}